\documentclass[11pt,a4paper]
{article}

\usepackage{graphicx}
\usepackage{amsmath}
\usepackage{amsfonts}
\usepackage{amssymb}
\usepackage{enumerate}
\usepackage{lscape}
\usepackage{longtable}
\usepackage{rotating}
\usepackage{multirow}
\usepackage{color}
\usepackage{url}
\usepackage{bm}
\usepackage{subfigure}

\usepackage{rotating}
\newtheorem{theorem}{Theorem}[section]

\usepackage[ruled,vlined,noline,linesnumbered]{algorithm2e}

\newtheorem{
claim}{Claim}[section]

\newtheorem{definition}{Definition}[section]

\newtheorem{assumption}{Assumption}[section]

\title{Learning Dynamic Bayesian Networks from Data: Foundations, First Principles and Numerical Comparisons}
\author{Vyacheslav Kungurtsev, Fadwa Idlahcen, Petr Ry{\v s}av{\' y}, Pavel Ryt{\' i}{\v r}, Ale{\v s} Wodecki}
\date{June 2024}

\begin{document}

\maketitle

\begin{abstract}
In this paper, we present a guide to the foundations of learning Dynamic Bayesian Networks (DBNs) from data in the form of multiple samples of trajectories for some length of time. We present the formalism for a generic as well as a set of common types of DBNs for particular variable distributions. We present the analytical form of the models, with a comprehensive discussion on the interdependence between structure and weights in a DBN model and their implications for learning. We give the analytical form of the likelihood and Bayesian score functions, emphasizing the distinction from the static case. We discuss functions used in optimization to enforce structural requirements. Next, we give a broad overview of learning methods and describe and categorize them based on the most important statistical features, and how they treat the interplay between learning structure and weights. We briefly discuss more complex extensions and representations. Finally we present a set of comparisons in different settings for various distinct but representative algorithms across the variants. 
\end{abstract}

\section{Introduction}
In this paper we give a comprehensive presentation on the training of Dynamic Bayesian Networks (DBNs), including both structure and parameters, from data. DBNs present a naturally interpretable model when it comes to understanding the precise interaction underlying the relationship between random variables. That is, the conditional independence structure defined by the DBN provides information regarding the mechanistic procedure that defines the model. This is also associated with the field of statistics referred to as causal learning. 

There is one general survey article on DBNs we found is~\cite{shiguihara2021dynamic}, which provides a helpful comprehensive resource for references for DBN modeling, inference and learning. In this work rather than seeking to provide a comprehensive literature review, instead we focus on narrating the global landscape of the mathematical understanding of the most important considerations as far as learning a model, including both the structure and parameters, from data.

With this understanding we are able to establish an informative taxonomy regarding methods, providing transparency in regards to the function and intention of each approach. There are important subtle distinctions as far as modeling assumptions between some different popular methods, and their awareness is critical for best practices of DBN learning. Finally, we perform comprehensive numerical comparisons, highlighting the particular advantages and disadvantages of each method. We highlight that these comparisons are not meant to be exhaustive or authoritative, but more informative and illustrative in regards to the tradeoffs associated with learning DBNs from data. 

We make several assumptions for this work. These are not entirely formal for the sake of precise Theorem proving, but rather a general restriction of the data generating process of interest, so as to highlight the most pedagogical features of DBN learning, as far as it is most commonly done in the literature. These assumptions are restrictive as far as faithful statistical modeling of real world phenomena. However, they tremendously simplify the learning process, and thus allow a more comprehensible presentation of what can one hope to achieve with standard simple methods.  For transparency, we present them below:
\begin{enumerate}
    \item \emph{Causal Sufficiency} There are no unobserved confounders, i.e., there is a closed system defining the data generating process wherein all causal sources are observed. This permits for the conditional independence structures to be reliable indicators of edge links in the graph. This is a standard formal assumption in almost all learning algorithms for DBNs and BNs.
    
    \item \emph{Causal Identifiability} we focus on the general case wherein the data regime permits for potential identification of a true causal graph, generally corresponding to the number of data samples (trajectories $\times$ time steps) some exponential factor of mangitude greater than the number of variables (in practice this can mean $5$ variables, $100$ trajectories of each $50$ time steps as a generic example). This permits us to focus on integer programming, constraint based methods, and other techniques that can obtain statistically significant point estimates for exact structures that recover a ground truth. This assumption is standard in the literature of DBN and SEM learning, as modeling uncertainty in less favorable data regime circumstances presents significant methodological challenges and considerations that require significantly more advanced techniques. However, understanding the nuances of the foundations are essential as far the proper development, implementation, and use of these techniques.
    
    \item \emph{Fully Observed} There are no hidden variables, all quantities of interest are fully observed at every time step. Of course, graphs with hidden (latent) variables and entire structures are instrumental for modeling in many fields. However, the inclusion of latent variables and the required Expectation Maximization modification of the procedures described presents technical complications that add significant additional complexity, and thus would necessity a much greater length while obfuscating the message of the fundamentals.
\end{enumerate}
We make a few departures from these assumptions throughout the work, which we explicitly indicate.

\subsection{Contribution of this Work}

In this work we present:
\begin{enumerate}
    \item A thorough explicit analytical description of standard popular DBN representations and statistical models. This includes the structure of potential dependencies of transition functions of the time-dependent random variables on other variables, time independent as well as time-dependent and in-time, Markovian one time step back, and delayed dependence. 
    \item Extensive commentary and analysis of learning DBNs from data from both the classical PGM/BN perspective as well as the time series perspective. The relationship of learning to the structure of the data, as well as high level intuition on the complex interaction of learning the structure and the parameters is presented.
    \item A presentation of the most standard and common criteria for defining the objective or cost of a particular DBN network structure as well as the functional forms of equations that define that the graph satisfies the structural requirements of a DBN, in particular acyclicity 
    \item Numerical results for examples of popular algorithms for learning DBNs, evaluated on a range of criteria and variety of problems. The set of examples is not meant to be exhaustive, nor the comparisons authoritative, but broadly illustrative of the relative advantages and disadvantages of the different methods available.
\end{enumerate}

\subsection{Applications of Dynamic Bayesian Networks}
The discovery of dynamic Bayesian networks has found many applications, including medicine \cite{Collett23, Eldawlatly08, lady22, Bueno16}, economics \cite{Ling15, LIU201946} and aviation \cite{Matthews13, Valdes18, gomez18}. The applications related to aviation typically involve finding casual structure in a sub-problem on the dispatch of flights and focused on risk mitigation. The medical applications typically focus on either the discovery of fundamental principles concerning chemical reactions taking place in biological organisms or extracted clinical data. In economics, it is typically of interest to uncover relationships between the stock market and macroeconomic and other market indices that either influence or are influenced by it.
In the following, we give some details about chosen applications as well as specific outcomes that modelling using DBNs has had in practice.

\subsubsection{Medical Science}
To highlight the importance of DBN-based discovery in medicine, we detail three separate applications. The first of these focuses on the quantification of disease development \cite{Bueno16}. Understanding the progression of diseases is crucial in clinical medicine, as it informs the effectiveness of treatments. Most clinical medicine and pathology textbooks provide detailed descriptions of disease progression and treatment responses. However, there has been limited research quantifying these descriptions in detail. Typically, research examines the temporal aspect by describing treatment outcomes after a certain period. A significant challenge in gaining deeper insights is the relatively small size of clinical datasets, often comprising only a few hundred patients.

In the aforementioned contribution, a heuristic procedure is proposed for exploring and learning non-homogeneous time dynamic Bayesian networks, aiming to balance specificity and simplicity. The approach begins with a fully homogeneous (in time) model parts of which are gradually replaced with sub-models which reflect the expected structure change at a given level of time delay. Furthermore, a splitting technique is applied to further improve the predictive behavior of the model, such models are typically termed partitioned DBNs.

A heuristic method was proposed to learn the DBN on synthetic data, which has a structure that should reflect a real data set. The numerical performance in terms of accuracy and solution time reported is hopeful. However, the method has yet to be tested on real world datasets.

The second application is concerned with the mapping of neural pathways \cite{Eldawlatly08}. Identifying functional connectivity from simultaneously recorded spike trains is crucial for understanding how the brain processes information and instructs the body to perform complex tasks. The study investigates the applicability of dynamic Bayesian networks (DBNs) to infer the structure of neural circuits from observed spike trains. A probabilistic point process model was employed to assess performance. The results confirm the utility of DBNs in inferring functional connectivity as well as the directions of signal flow in cortical network models. Additionally, the findings demonstrate that DBNs outperform Granger causality when applied to systems with highly non-linear synaptic integration mechanisms.

The third chosen application focuses on the choice of appropriate treatment regimens for Chronic lymphocytic leukemia (CLL) \cite{lady22}. This cancer is the most common blood cancer in adults, with a varied course and response to treatment among patients. This variability complicates the selection of the most appropriate treatment regimen and the prediction of disease progression. The aforementioned paper aims to develop and validate dynamic Bayesian networks (DBNs) to predict changes in the health status of patients with CLL and predict the progression of the disease over time. Two DBNs, the Health Status Network (HSN) and the Treatment Effect Network (TEN), were developed and implemented. Relationships linking the most important factors influencing health status and treatment effects in CLL patients were identified based on literature data and expert knowledge. The developed networks, particularly TEN, were able to predict the probability of survival in CLL patients, aligning with survival data collected in large medical registries. The networks can tailor predictions by integrating prior knowledge specific to an individual CLL patient. The proposed approach is a suitable foundation for developing artificial intelligence systems that assist in selecting treatments, thereby positively influencing the chances of survival for CLL patients.

\subsubsection{Economics}
The relationship between the stock market and national economies deepens as the market matures, highlighting the need to study their dynamic interplay. Economic indicators such as real income and savings rates play crucial roles in influencing stock market capitalization. Macroeconomic fundamentals wield considerable influence over both short and long-term periods. Some researchers argue that finance and economic growth are causally linked, suggesting the stock market's potential to drive economic development. However, not all macroeconomic factors significantly impact stock prices. Understanding the strength of association among these variables offers insights into how the stock market behaves across varying economic landscapes. Research on the Chinese stock market examines how macroeconomic variables shape stock market indices over time, emphasizing the enduring influence of economic fundamentals amid short-term market fluctuations. The aforementioned interplay may be modeled by DBNs and has been detailed in \cite{LIU201946}.

In the article and analysis of the relationship between the stock market and economic fundamentals using 11 selected factors is modeled using a DBN. Among these, four factors pertain to stock market indicators, while the remaining factors focus on macroeconomic and policy considerations. The first four factors reflect stock market performance, with defensive and cyclical stocks exhibiting varying behaviors during bull and bear markets. The Stock Exchange 50 index, comprising the 50 largest and most liquid stocks in the Shanghai Securities Market, supplements the overall stock market observation. Additionally, the consumer stock index serves as an indicator of societal consumption levels, typically rising during favorable macroeconomic periods and declining during economic downturns.

The results on real data of the described method are mixed. The application to the Shanghai composite market yielded some positive results in terms of the prediction of macroscopic quantities, but only limited success in terms of constituent market price prediction. The modeling of the components of the market in sufficient detail is a difficult problem due to the numerical tractability being limited as the number of variables increases. 

\subsubsection{Aviation}
The global collection of aircraft and the airspace in which they operate is a complex system generating a vast amount of data, making it a challenging domain to model mathematically. This system includes critical elements such as aircraft, airports, flight crews, weather events, and routes, each with many subcomponents. For example, aircraft have numerous subsystems and components, each subjected to various stresses and maintenance actions, which influence their time dynamics. Airports have multiple runways and internal logistical processes, which influence the operational capacity. These system components interact in complex ways. For instance, each flight corresponds to an aircraft operated by a flight crew traveling from one airport to another via a route that may need to change due to weather. Multiple flights operate simultaneously, requiring coordination to avoid incidents while maximizing throughput and minimizing delays.

To give an idea about a specific aviation problem that may be tackled with DBNs, we describe the airport operation uncertainty characterisation developed in \cite{gomez18}. The model outlines aircraft flow through the airport, emphasizing integrated airspace and airside operations. It characterizes various operational milestones based on an aircraft flow’s Business Process Model and Airport Collaborative Decision-Making methodology. Probability distributions for factors influencing aircraft processes need to be estimated, along with their conditional probability relationships. This approach results in a dynamic Bayesian network that manages uncertainties in aircraft operating times at the airport. The nodes of the network describe various aspects of the airport and flight operations. They cover meteorological conditions, arrival airspace variables such as timestamps and congestion metrics, airport infrastructure, operator and flight data, airside operational times and flight regulations, and the causes of delays. 

The key outcomes of this work include the statistical characterization of processes and uncertainty drivers, and a causal model for uncertainty management using a DBN. Analyzing 34,000 aircraft operations at Madrid Airport revealed that arrival delays accumulate throughout the day due to network effects, while departure delays do not follow this pattern. The major delay drivers identified were the time of day, ASMA congestion, weather conditions, arrival delay amount, process duration, runway configuration, airline business model, handling agent, aircraft type, route origin/destination, and ATFCM regulations. Departure delays are significantly impacted by events of longer duration, which also offer greater potential for recovery.

\section{Background - Dynamic Bayesian Networks}
We present the general, and then specific forms, of DBN models. Consider that there is an $n_x$-dimensional stochastic process $X(t)$. The individual random variables $X_i(t)$ for all $i\in[n_x]$ can be valued as discrete,  or as members of some field, such as $\mathbb{R}$. In addition, there can be an $n_z$-dimensional (time independent) random variable $Z$. Let us denote the generic spaces as $\mathcal{X}$ and $\mathcal{Z}$, respectively.

The defining character of DBNs is modeling the dependence of $X_i(t)$, for each $i$, on other quantities, i.e., defining the the evolution of the stochastic process $X(t)\to X(t+1)$. Formally, for the probability kernel defining the iterations of the stochastic process, the dependence must be Markovian, with additional time lagged auto-regressive effects, that is
\begin{equation}\label{eq:dbndefn}
p(X_i(t+1)\in A) = f_i(X(t),X_{j\neq i}(t+1),\{X_i(t-\tau)\}_{\tau=1,...,p},Z)
\end{equation}
where $A$ is some set in the Borel $\sigma$-algebra of $\mathcal{X}$. That is, the transition kernel can depend on the current state of the other random variables, the values of the random variables at the previous time, the time-independent variables $Z$, as well as a possibly autoregressive effect through dependence on $\{X_i(t-\tau)\}_{\tau=1,...,p}$. 

In addition, there is the important requirement that no in-time string of dependencies, that is from $X_{j\neq i}(t+1)$ to $X_i(t+1)$, forms a cycle. This presents the necessity of introducing graph theoretical notions to precisely characterize DBNs. Generically, we say a directed graph is a set of vertices $\mathcal{V}=\{v_1,v_2,...,v_n\}$ and edges $\mathcal{E}=\{e_1,e_2,...,e_m\}$, where $e_j=(v_{j_1},v_{j_2})$ denotes the existence of a directed path between the two notes $v_{j_1}\to v_{j_2}$. We also say in this case that $j_1$ is a parent of $j_2$, or $j_1\in dpa(j_2)$.

The DBN model structure is defined by a graph superset, 
\begin{equation}\label{eq:fullgraph}
\bar{\mathcal{G}}=\mathcal{G}(\mathcal{V}(X(t-1),X(t)),\mathcal{E}_d)\cup\mathcal{G}(\mathcal{V}(X(t),X(t)),\mathcal{E}_s)\cup \mathcal{G}(\mathcal{V}(X(t),Z),\mathcal{E}_z)\cup \mathcal{G}(\mathcal{V}(X(t),X(t-\tau)),\mathcal{E}_{\tau})
\end{equation}
The first two union operands define connections in the model between the temporal random variables. That is $e=\{V_1,V_2\}\in\mathcal{E}_d$ with $V_i\in \{\{X_i\}\}$ if the transition for $V_2=X_i(t+1)$, that is $p(X_i(t+1)\in A)$, is a function of $V_1=X_j$, or in other words $j\in dpa_d(i)$. For the static dependencies  $j\in dpa_s(i)$, and with $V_1=X_i$ and $V_2=X_j$, it holds that $e=\{V_1,V_2\}\in\mathcal{E}_s$ if $p(X_j(t+1)\in A)$ is a function of $V_1=X_i(t+1)$. Finally, we also have a (non-symmetric) matrix encoding the dependencies on the self-history $\tau\in dpa_{\tau}(i)\subset \{1,...,p\}$ and the dependencies on the static variables $dpa_z(i) = \left\{Z_j: \frac{\partial f_i(\cdot)}{\partial Z_j}\neq 0 \right\}$. These, of course, can be encoded as graphs as well. Generically, the autoregressive effects can include lagged influence across distinct random variables, however for simplicity we focus only on including self-dependencies for time lags longer than one.

This permits us to write~\eqref{eq:dbndefn} as,
\begin{equation}\label{eq:dbndefncompact}
p(X_i(t+1)\in A) = f(A,\{X_j(t)\}_{j\in dpa_d(i)},\{X_j(t+1)\}_{j\in dpa_s(i)},\{X_i(t-\tau)\}_{\tau\in dpa_{\tau}(i)},\{Z_j\}_{j\in dpa_z(i)})
\end{equation}
Notice that the encoding of the explicit dependencies presents the possibility of using a common $f$ as opposed to one depending on the transition out-node $i$, in the case wherein all the variables $X_i$ are of the same distributional family. This eases the computation of the likelihood of the data given the parameters and structure, etc. 

We will sometimes use, for shorthand:
\begin{equation}\label{eq:defvdpa}
    \{V_j(t+1)\}_{j\in dpa(i)}=\{X_j(t)\}_{j\in dpa_d(i)}\cup\{X_j(t+1)\}_{j\in dpa_s(i)}\cup\{X_i(t-\tau)\}_{\tau\in dpa_{\tau}(i)}\cup\{Z_j\}_{j\in dpa_z(i)}
\end{equation}

\begin{figure}
\begin{center}
    \includegraphics[scale=0.33]{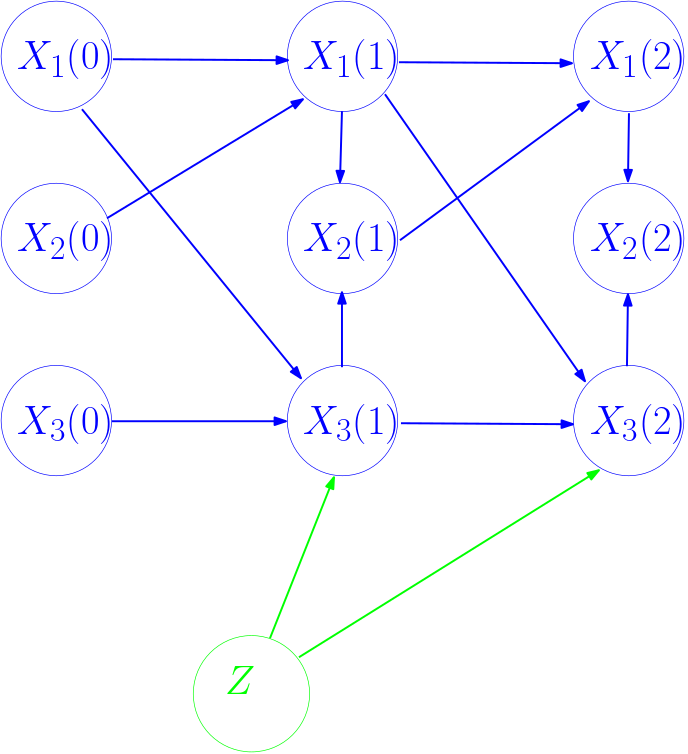}
    \end{center}
    \caption{A Possibl DBN Graphical Network defining the transitions of $X(t)$}\label{fig:dbnpic}
\end{figure}

See Figure~\ref{fig:dbnpic} for an illustration of a DBN and its transition dynamics. In this case for $i=1$, the Markovian transitions are from itself and from $X_2$, and there are no intra time nodes or static nodes directed to it, and so $V_{dpa(1)}(t)=V_{dpa_d(1)}=\{X_1(t-1),X_2(t-1)\}$. For node 2, there are no Markovian transitions and two intra-node dependencies, thus $V_{dpa(2)}(t)=V_{dpa_s(2)}(t)=\{X_1(t),X_3(t)\}$. Finally, for $X_3(t)$, there are two Markovian dependencies, and a static covariate dependence. Thus $V_{dpa(3)}(t)=V_{dpa_d(3)\cup dpa_z(3)}(t)=\{X_1(t-1),X_3(t-1),Z\}$.

Now that we have established the general form of the DBN, we see that we have a fundamentally still very general problem to solve, in that the function $f$ can encode any sort of dependency on the different variables in the parent set of the note of interest. They can depend as according to various nonlinear interactions, that can themselves embody different conditional independence information In order to complete the model, we need to define the form of the function $f$.

\subsection{Simple Parametric Conditional Probability Dependency}
For certain kinds of variables, it becomes both possible and prudent to use certain simple parametric families for defining $f$. For instance, for binary Bernoulli random variables, the use of Dirichlet distributions for the prior of the weights together with using Conditional Probability Tables to define the form. For continuous random variables, Gaussian linear models provide a means of computing the maximum likelihood linear parameters using covariance matrices. 

A significant advantage of using parametric families arises from the closed form computation of criteria, which permits closed form computation of the marginal posterior of a structure. This permits structure learning algorithms to be able to score graphs offline, assisting in the search. Many score maximizing procedures such as~\cite{geiger2002parameter,cussens2020gobnilp,BARTLETT2017258,atanackovic2024dyngfn} use this approach. The score is ultimately an integration of the posterior of the parameters in the model given the structure, which also indicates that the sampling of the optimal parameters, once obtaining the maximum a posteriori structure, is straightforward for these models. 

With the optimal parameters and likelihood computation amortized, one can then use black box approaches such as those involving neural models. This includes the innovative Generative Flow Network approaches as given by~\cite{pmlr-v180-deleu22a,atanackovic2024dyngfn}. These use a Reinforcement Learning iteration to ultimately sample from a high scoring network as according to a defined score. Reinforcement Learning broadly, e.g.,~\cite{zheng2023rbnets} is another framework by which the structure search for these standard specific models can be aided by neural networks. 

Linear Structural Equation Models (SEMs) present an opportunity to use an adjacency matrix to define both the structure and weights in a computationally advantageous form. This highlights the correspondence between the Dynamic Systems and the graph theoretic developments in causal learning.

\subsubsection{Discrete Variables}

\paragraph{Binary Variables}
The case of binary random variables is wherein $X_i(t)\in B(1,p^X_i(t)),Z_j\sim B(1,p^Z_j)$, etc., that is, they are all of Bernoulli type. Empirical samples for all $k\in[K]$, where $k$ indexes a set of sample trajectories satisfy $X^{(k)}_i(t),Z^{(k)}_j\in \{0,1\}$ for all $t$, $i$, and $j$. With this most simple scenario, the modeling flexibility as well as the nuances of structure learning becomes a natural pedagogical start. 

There is a degree of flexibility in the choice of statistical model for defining the transition function. We will explore three options - the noisy or model, the linear logit model, and the complete dependency table model. 

The \emph{noisy or model} defines the transitions as 
\begin{equation}\label{eq:noisyor}
\begin{array}{l}
p(X_i(t+1)=0) = (1-\lambda_0)\prod_{l\in dpa(i)} (1-\lambda_l)^{V_l} \\
p(X_i(t+1)=1) = 1-(1-\lambda_0)\prod_{l\in dpa(i)} (1-\lambda_l)^{V_l} 
\end{array}
\end{equation}
This model is referred to as noisy or because essentially it calculates a probabilistic perturbation of the binary OR operation. This model presents one implementation of causal independence, wherein the influence of each covariate is independent with respect to the others.

For the linear logit models, define the sigmoid function, 
\[
\sigma(x) = \frac{e^{x}}{1+e^x}
\]

The reason the models we define next are referred to as linear is that the transition is defined to be,
\begin{equation}\label{eq:binarylinear}
X_i(t+1) = \beta_0+ \sum\limits_{j\in dpa_d(i)} \beta^d_j X_j(t)+
\sum\limits_{j\in dpa_s(i)} \beta^s_j X_j(t+1) +\sum\limits_{\tau\in dpa_{\tau}(i)} \beta^a_{\tau} X_i(t-\tau)+\sum\limits_{j\in dpa_z(i)} \beta^z_j Z_j
\end{equation}
We shall see that this linear form is broadly common in modeling the transitions of variables in DBN models for other variable types. 

The probability kernel given by~\eqref{eq:binarylinear} is
\begin{equation}\label{eq:binarylinearkernel}
p(X_i(t+1)=1) = \sigma\left(X_i(t+1)\right)
\end{equation}
with $X_i(t+1)$ given by the expression above.

One alternative that frequently arises in practice is the necessity to accurately model \emph{Conditional Probability Dependencies} (CPDs) as defined by \emph{Conditional Probability Tables} (CPTs). As an example, please see Table~\ref{tab:cpt}. 

\begin{center}
\begin{table}\caption{An example of a CPT}\label{tab:cpt}
\begin{center}    \begin{tabular}{|c c c | c |}
    \hline
    $X_2(t)$ & $X_3(t+1)$ & $Z_2$ & $X_1(t+1)$ \\ \hline
    0 & 0 & 0 & 0 \\
    0 & 0 & 1 & 1 \\
    0 & 1 & 0 & 1 \\
    0 & 1 & 1 & 0 \\
    1 & 0 & 0 & 1 \\
    1 & 0 & 1 & 0 \\
    1 & 1 & 0 & 0 \\
    1 & 1 & 1 & 1 \\ \hline 
    \end{tabular}\end{center}
    \end{table}
\end{center}
It is clear that the information in Table~\ref{tab:cpt} cannot be modeled with a linear transition function as in~\eqref{eq:binarylinear}. In this case, if one wanted to construct such a model, one would instead have to be able to include all of the combinations between the possible parent nodes. 

Formally, a transition model could look like, for Table~\ref{tab:cpt},
\[
\begin{array}{l}
X_1(t+1) = \beta_0+\beta_{1}Z_2+\beta_2 X_3(t+1) +\beta_3 X_3(t+1) Z_2+\beta_4 (1-X_3(t+1)Z_2)+\beta_5 X_2(t) \\ \qquad\qquad\qquad +\beta_6(1- X_2(t) Z_2) +\beta_7 (1-X_2(t) X_3(t+1))+\beta_8 X_2(t) X_3(t+1) Z_2 
\end{array}
\]
and in the general case,
\begin{equation}\label{eq:cpd}
\begin{array}{l}
X_i(t+1) = 
\prod\limits_{j\in
dpa_d(i)} \prod\limits_{k\in dpa_s(i)}\prod\limits_{\tau=1,...,p}\prod\limits_{l\in dpa_z(i)} \sum_{\alpha\in\left(\mathbb{Z}^+_2\right)^4} \beta_{i,j,k,l,\tau}^\alpha\left(W_{jkil}\right)\\
W_{jkil} = \Pi_{jkil} \left(X_j(t) X_k(t+1)X_i(t-\tau) Z_j\right)^{\alpha}\\ \qquad\qquad +\left(1-\Pi_{jkil}\right)\left(1-\left( X_j(t) X_k(t+1)X_i(t-\tau) Z_j\right)^{\alpha}\right),\,\Pi_{jkil} \in \{0,1\}
\end{array}
\end{equation}
where the parameters are appropriately normalized. With the combinatorial explosion in this model clearly visible, it can be seen that such circumstances present significant difficulties as far as computing hardware expense in both processing and memory, when it comes to modeling high dimensional datasets. 

On the other hand, this structure of statistical model presents two structural conditions denoted as \emph{local parameter independence} and \emph{unrestricted multinomial distribution} \cite{spiegelhalter1990sequential}. These ensure that for every \emph{configuration}, that is, every possible combinations of values of the parents of a node, there is an independent parameter vector. This leads to a corresponding combinatorial explosion of parameter vectors in the statistical model. On the other hand, however, the analytical calculation of parameter likelihoods and posterior distributions become possible, facilitating more straightforward evaluation of scoring metrics quantifying the information quality of an entire (D)BN. That is, the marginal likelihood of a structure can be computed without first computing the likelihood of the weights.

\paragraph{Multinomial}
Multinomial distributions are over discrete valued random variables that can take on multiple possible values. The distinction between a user friendly linear parameter presentation and the expressiveness at the cost of parametric dimensionality of unrestricted multinomial distributions becomes apparent in the increased complexity of modeling multinomial relative to Bernoulli distributions. 

Now, consider that for every $i$, $X_i(t)\in \{u^1,...,u^m\}$ some multinomial sample, with Dirichlet sampled initial values, and always with some multinomial distribution $\{\theta_i^m(t)\}$. The set of parameters indicating the probability that $X_i(t+1)=u^k$ given a particular configuration of the parent nodes $V_{dpa(i)}(t+1)$ is denoted $\theta_{i,v_{dpa(i)}}^k$. 

First, consider a linear model. Let us simplify the notation,
\[
\sum_{j\in dpa(i)} \beta_j V_j(t+1) = \sum\limits_{j\in dpa_d(i)} \beta^d_j X_j(t)+
\sum\limits_{j\in dpa_s(i)} \beta^s_j X_j(t+1) +\sum\limits_{\tau\in dpa_{\tau}(i)} \beta^a_{\tau} X_i(t-\tau)+\sum\limits_{j\in dpa_z(i)} \beta^z_j Z_j 
\]
With this,
the form of the transition probability is,
\begin{equation}\label{eq:multinom}
p(X_i(t+1)=u^l) = \frac{\exp\left(\beta_{i,0}+\sum_{j\in dpa(i)} \sum_{q\in[m]}\beta^{l,q}_{i,j} \mathbf{1}(V_j=u^q)\right)}{\sum_{s\in[m]}
\exp\left(\beta_{i,0}+\sum_{j\in dpa(i)} \sum_{q\in[m]}\beta^{s,q}_{i,j} \mathbf{1}(V_j=u^q)\right)}
\end{equation}
which is a standard linear logit.

On the other hand, with an unrestricted multinomial distribution, we can define the full transition distribution explicitly, meaning for every possible combination of values instantiated by a node's parents in a given network, we define a specific probability. In this case even the already cumbersome notation of~\eqref{eq:cpd} is insufficient to present the model representation. On the other hand, we will see that this representation eases the likelihood and Bayesian score computations. Finally, we distinguish $dpa(i)=dpa_t(i)\cup dpa_z(i)$ as the time-dependent and time-independent covariates. We also distinguish the possible values of $Z$ to be $Z_j\in \{1,...,w^q\}$

We simply denote: 
\begin{equation}\label{eq:genericthetaxi}
p(X_i(t+1)=u^l\vert V_{dpa(i)}(t+1),\theta_i) := \theta_i^{\xi(V_{dpa(i)}(t+1))},\,\xi\in \Xi_i,\,
\Xi_i= \prod_{j\in dpa_t(i)} \mathbb{Z}_{m}^+ \times \prod_{j\in dpa_z(i)} \mathbb{Z}_q^+
\end{equation}
That is, there is an multi-index that enumerates the entries of $\Xi_i$ for each transition $i$. We can see that this presents a highly parametrized model, which will imply significant parametric uncertainty when there are finite data samples. On the other hand, with this highly precise model, the maximum likelihood becomes much more straightforward to compute, as well as the Bayesian scores. Indeed, this is exactly what local parameter independence facilitates -- you can compute the likelihood by counting the instances of each transition and dividing by the count of each predecessor configuration. On the other hand, when the total count of every possible predecessor configuration is low, due to unfavorable sample complexity, this can be a rather noisy source of actual validity of that dependence. On the other hand, by letting these parameters take on distributions, in a Bayesian setting, the computation of a posterior for a structure becomes easier, and the uncertainty is available by sampling the posterior, anyway. 

\subsubsection{Continuous Variables}

\paragraph{Gaussian Variables}

A Gaussian Bayesian Network can be considered a continuous variable equivalent to binary variables in the sense that the structure permits closed form expressions for computing the likelihood, posterior, etc. In this case, however, the additive linear term is standard. Formally, we assume that $X(t)\sim\mathcal{N}(\mu;\Sigma)$. The transition function becomes:
\begin{equation}\label{eq:gausslinear}
\begin{array}{l}
p(X_i(t+1) \vert dpa_d(i)\cup dpa_s(i)\cup dpa_{\tau}(i)\cup  dpa_{z}(i)) = \mathcal{N}\left(\beta_0+\sum\limits_{j\in dpa_d(i)} \beta^d_j X_j(t) \right. \\ \qquad\qquad \left. +
\sum\limits_{j\in dpa_s(i)} \beta^s_j X_j(t+1)+\sum\limits_{\tau\in dpa_{\tau}(i)\subset \{1,...,p\}} \beta^a_{\tau} X_i(t-\tau)+\sum\limits_{j\in dpa_z(i)} \beta^z_j Z_j; \sigma^2\right)
\end{array}
\end{equation}
and the result is that,
\[
\begin{array}{l}
\mu_{X(t+1)} = \beta_0+\beta^T \mu ,\, \sigma^2_{X(t+1)}=\sigma^2+\beta^T \Sigma \beta, \\ 
Cov\left[\{X(t)\},\{X(t+1)\},\{X(t-\tau)\},Z;X(t+1)\right] = \sum \beta_j \Sigma_{i,j}
\end{array}
\]
Indeed in~\cite[Theorem 7.3-7.4]{koller2009probabilistic} it is shown that there is a bidirectional equivalence between such a normal joint distribution and normal transition function. 

We shall see that this permits, in the temporal case, a repeated composition of the propagation of the covariance with each time step, when computing the likelihood and performing inference. This is associated with the deep theory of filtering methods, which typically studies Gaussian DBN propagation with a simple state-observable structure.

\paragraph{Exponential Family Functional Form}

An exponential family is defined with, recalling $\mathcal{X}$ to be an abstract space for which both $X(t),Z\in \mathcal{X}$,
\begin{enumerate}
    \item A sufficient statistics function $\tau:\mathcal{X}\to \mathbb{R}^K$ for some $K$
    \item A convex set of a parameter space $\Theta\subset \mathbb{R}^m$
    \item A natural parameter function $t:\mathbb{R}^m\to \mathbb{R}^K$
    \item A measure $A$ over $\mathcal{X}$
\end{enumerate}
The exponential family is a distribution of the form
\begin{equation}\label{eq:expfam}
P_{\theta}(\xi) = \frac{1}{Z(\theta)} A(\xi)\exp\left\{(t(\theta),\tau(\xi)\right\},\,Z(\theta)=\sum\limits_{\xi} A(\xi) \exp\left\{(t(\theta),\tau(\xi))\right\}
\end{equation}

The case of natural parameters is the most standard, and the one we have been exploring in the formulations above, this corresponds to $(t(\theta),\tau(\xi))=(\theta,\tau(\xi))$. One has to be careful, however, in constraining the space of parameters $\theta$ to ones normalized, i.e.,
\[
\Theta =\{\theta\in \mathbb{R}^m:\int \exp((\theta,\tau(\xi)))d\xi < \infty \}
\]

\paragraph{Linear Structural Equation Models}
Consider the general case wherein the function $f$, as given by~\eqref{eq:dbndefn}, is given a linear parametrization with respect to continuous variables $X(t),Z$, as in~\eqref{eq:binarylinear}, however for continuous variables. One can then perform learning by minimizing the appropriate least squares fit to the data. This is most common in the approach of Linear Structural Equation Models, in which case a linear parametrization permits greater computational ease. 

Linear Structural Equation Models (LSEMs) are the most common non-Gaussian DBN for modeling continuous variables. With LSEMs (see, e.g. \cite{bowen2011structural} for a general reference and~\cite{peters2014causal} for application to causal inference) presume a general linear structure that is associated with a discretization of a dynamical system:
\[
\dot{X}(t) = f(X(t),Z)
\]
with this generality, there is a degree of ambiguity in the literature, because there are a number of ways to consider a discrete model of this. 

An SEM could refer to a purely time-instant (static) model, with dependencies $dpa_s$ and $dpa_z$ only, as in~\cite{pmlr-v180-deleu22a,manzour2021integer}. More recently, DBNs more broadly have become interchangeable with SEMs, for instance, the representation in
\cite{pamfil2020dynotears} has all dependencies as described here except, it can be argued for simplicity, $Z$.


\paragraph{Nonlinear, Nonparametric and Neural Models}
The structure of $f$, or even if there is an $f$ at all, is of course flexible like with any statistical modeling. More complex statistical models for the transition introduce significant additional difficulties in training, by adding nonconvexity to the landscape and significantly expanding the degrees of freedom in the model that need to be fit with data. Given the emphasis in this article on simple illustrative DBNs, we will but briefly mention some examples, neither comprehensive nor authoritative.

Broadly speaking, there are a number of popular parametric forms of nonlinear models that can be used from time series literature, e.g.~\cite{douc2014nonlinear}. Neural networks have enabled computationally intensive empirical unsupervised time series models 
\cite{franceschi2019unsupervised,cai2024msgnet}. Simple nonlinear functional models in the SEM statistical community have also been studied~\cite{lee2000statistical}. The work~\cite{sharma2022scalable} uses splines to model the nonlinear relationships in the transition distributions. The work~\cite{kim2004dynamic} uses a kernel nonparametric regression model to learn DBNs for gene regulatory networks.

\section{Learning From Multiple Trajectories}\label{s:learningfromtraj}
Consider that we receive $N$ samples of trajectories $\mathcal{T}^j$, each with a total time of $T$ 
\begin{equation}\label{eq:sample}
\mathcal{S}=\cup_{n=1}^N \mathcal{T}^n=\{ Z^{(n)},X^{(n)}(0),X^{(n)}(1), X^{(n)}(2), ..., X^{(n)}(T)\}_{n=1,...,N}
\end{equation}
and we are interested in fitting a DBN model to this data. This amounts to defining the specific form of $f$ in~\eqref{eq:dbndefn} and as far as three tasks, as defined in the introduction. More specifically, it amounts to 
\begin{enumerate}
    \item identifying the parents of each $X_j(t)$ in the graph $\bar{\mathcal{G}}$, as well as specific functional form of the transition function $f$, of a ground truth
    \item Performing inference on the model accurately. For DBNs a common use is for forecasting, but generalization and test error is another common metric.
\end{enumerate}

\subsection{Maximum Likelihood Calculations}
In reviewing the literature on learning DBNs from data, it is typical to disregard the distinction of the trajectory sample $\mathcal{T}^i$ and the time transition samples $\{X^{(i)}(t),X^{(i)}(t+1)\}$. As far as understanding the meta-methodological cause of this, it appears that this can be said to be due to DBNs being considered not uniquely, but as either a special kind of Bayesian Network, or as splices of the same time series trajectory. 

Consider the two methodological components surrounding the study of DBNs, time series analysis and PGMs. For the latter, consider two popular works that are effectively extensions of methods developed for BNs extended to DBNs, the continuous reformulation of the problem into one with adjacency matrices as decision variables, called ``NOTEARS'' in the static case~\cite{zheng2018dags} and ``dynotears''~\cite{pamfil2020dynotears}, as well as the use of ``Generative Flow Networks'', a Reinforcement Learning-motivated sampler, for the static case in~\cite{pmlr-v180-deleu22a} and the dynamic case in~\cite{atanackovic2024dyngfn}. It can be seen that in all of these cases, the likelihood is expressed as,
\begin{equation}\label{eq:likelihood}
\begin{array}{l}
    p\left(\mathcal{S}\vert \theta_G,\bar{\mathcal{G}} \right) = \prod\limits_{n=1}^N \prod\limits_{s=1}^{T}  p\left(X^{(n)}(T-s+1)\vert \theta,Z^{(n)},\{X_j^{(n)}(T-s)\},\right.\\\qquad\qquad\left.\{X^{(n)}_j(T-s+1)\},\{X^{(n)}(T-s-\tau)\}_{\tau=1,...,p}\right)
    \end{array}
\end{equation}
and with the standard application of the logarithm, change into a sum, and maximization, or a posteriori maximization through a Bayesian criterion, as the target. 

And similarly, in consulting standard texts on time series analysis with detailed derivations of Likelihood computation for various models, e.g.~\cite{douc2014nonlinear,mcquarrie1998regression}, we see that in the derivations of the likelihood, the data is considered to be a sequence of observations, that is, a sequence of observations from a stochastic process $\{\hat{X}(0),\hat{X}(1),\hat{X}(2),...,\hat{X}(T)\}$, rather than the general form given in~\eqref{eq:sample}, and is fit to~\eqref{eq:likelihood}, just with a simpler expression in the sum index. However, arithmetically, the calculations commute, that is, they are equivalent to the Bayesian Network paradigm. Being equivalent expressions for the likelihood, brought from different perspectives, we can see that the conditional nature of each sample makes the concern as to the appropriate application of the methods to be less as to whether they can adequately computed with sound statistical assumptions, but care as to what one can learn, i.e., learning conditional dependencies rather than long term statistical properties.

\subsection{Background on Causal Learning}
\paragraph{Conditional Independence and d(irected)-separation}
Let $G$ be a (D)BN.
Let $X_1, \dots, X_n$ be the set of random variables of (D)BN.
Let $V,W$ be subsets of $\{1,\dots, n\}$.
We say that the set $X_V$ is conditional independent of $X_W$ given $X_Z$ if the following condition holds:
$$P(X_V|X_W,X_Z)=P(X_V|X_Z).$$

Independence of various sets of variables can be determined by examining d-separation (d means directional) criterion of the (D)BN DAG \cite{ProbabilisticCowell}.

A (undirected) trail $T=(V_T,E_T)$ (path that does not contain any vertex twice) of $G$ is \emph{blocked} by the set $Z$ if $\forall v\in V(G)$ either (i) $v\in Z\cap V_T$, and in-degree of $v$ is at most 1; (ii) $v \notin Z$ and $\textit{children}(v)\cap Z =\emptyset$, and both arcs of $T$ connected to $v$ are directed to $v$. The sets $V$ and $W$ are d-separated if any trail between $V$ and $W$ is blocked by the set $Z$. If $V$ and $W$ are not d-separated, we say that they are d-connected.

The set $X_V$ is conditional independent of $X_W$ given $X_Z$, if $V$ and $W$ are d-separated by $Z$.

\paragraph{Causal Sufficiency}

We present the main set of definitions used in the literature from \cite{ ???, beckers_causal_2021}.

A signature \( S \) is defined as a tuple \( (U, V, R) \), where \( U \) is a set of exogenous variables, \( V \) is a set of endogenous variables, and \( R \) is a function that maps each variable \( Y \) in \( U \cup V \) to a nonempty set \( R(Y) \) of possible values, representing the range of values \( Y \) can take. For a vector of variables \( \vec{X} = (X_1, \dots, X_n) \), \( R(\vec{X}) \) represents the Cartesian product \( R(X_1) \times \dots \times R(X_n) \).

A causal model \( M \) consists of a pair \( (S, F) \), where \( S \) represents the signature and \( F \) specifies a set of functions. Each function in \( F \) associates an endogenous variable \( X \) with a structural equation \( F_X \), which determines the value of \( X \) based on the values of other endogenous and exogenous variables. Formally, the function \( F_X \) maps \( R(U \cup V - \{X\}) \) to \( R(X) \), meaning that \( F_X \) provides the value of \( X \) given the values of all other variables in the set \( U \cup V \).

\begin{definition}
 We say that $U=u$ is directly sufficient for $V=v$ if for all $c\in R(V-(X\cup Y)$ and all $u\in R(U)$ it holds that $(M,u)\models[X\leftarrow x, C\leftarrow c]Y=y$.   
\end{definition}

\begin{definition}
    We define that $X=x$ is strongly sufficient for $Y=y$ if there is an $N=n$ such that $Y\subseteq N$ and $y$ is a restriction of $n$ to $Y$ and $X=x$ is directly sufficient for $N=n$
\end{definition}

\begin{definition}
    We define $X=x$ is weakly sufficient for $Y=y$ in $M$ if for $u\in R(U)$ it holds that $(M,u)\models[X\leftarrow x]Y=y$
\end{definition}

\paragraph{Causal discovery and Inference}
The problem of Causal discovery is to find a true graph $G$ as the best possible explanation of the given data.
There are various causal discovery methods. Such as score-based algorithms, which try to recover the true causal graph by finding a graph that maximize a given scoring function. Another example are Constraints based algorithms or continuous optimization algorithms.

\subsection{Considerations from Axioms of Causal Learning}
DBNs, compared to BNs, contain both time-varying as well as static variables. While the in-time transition structure is still a DAG, suggesting that formally many of the same principles regarding inference as well as structure and weight learning in BNs carry over to DBNs, the presence of time, especially when long trajectories are expected, adds significant complications. 

Consider having a set of $T$ sampled trajectories. On the one hand each trajectory is i.i.d., but above that, each time point relative to the previous presents an additional sample, with additional information. This presents the question: how can we distinguish the amount, and specific utility, of information gained from an additional trajectory, versus that gained from an additional time point? 

This indicates the utility of including both static $Z$ and dynamic $X$ variables in the model for elucidation.  Instead one can consider a new trajectory as a new sample of $\hat{Z}$, which itself samples $X(0)\sim \pi(X(0)\vert Z)$ then, $X(0),X(1),...,X(T)$. As such, one has $T$ samples in order to learn $P\left(X(t+1)\vert X(t),\hat{Z}\right)$. However, what can be said about how informative a marginal trajectory is towards learning $P\left(X(t+1)\vert X(t)\right)$, that is, the marginal conditional over the population of $Z$?

It seems intuitive that in some way $P(\hat{Z})$ as well as $\pi(X(0)\vert Z)$ should weigh the in, where $P(\hat{Z})$ is the population prior of $P(\hat{Z})$, corresponds to the information gained for $P\left(X(t+1)\vert X(t)\right)$. For continuous variables, the information depends on the cross correlation as the prior evaluation is perturbed. It is clear then that information complexity is actually benefited from low variance, or low cardinality of a discrete space, between trajectories. Thus, the DBN model is particularly suitable for understanding long and complex time evolution of systems that do not change much in different contexts.

Recall that causal sufficiency requires that all confounding variables be present and observed. It is clear that different trajectories represent some distinctions in circumstance of object that the observations are taken from. If this is a latent variable, this presents an insurmountable probably to identification. 

As far the required observations for causal identification, this is that there exists at least one $Z$ such that for all trajectories $Z$ is observed, and $Z$ is in the parent of some $X(t)$. We can consider that the classic \emph{Randomized Clinical Trial} is exactly that $Z_H\in\{0,A\}$ and then testing for $p(X(t+1)\vert X(t),Z_{\setminus},Z_H=0)\neq p(X(t+1)\vert X(t),Z_{\setminus},Z_H=A)$, with a null and alternative hypothesis and $Z_{\setminus}$ as other covariates, assumed to me completely independent of $Z_H$.

The less that $Z_{\setminus}$ mediates the transitions, the more the trajectories can be treated as independent.


\subsection{Closed System Graph Causal Identification Model and Likelihood Information}
In an effort to establish appropriate first principles by which to study the computational and statistical properties of joint structure-parameter learning in DBNs, we will present two definitions of specific setting and problem. In this first case, we consider the more mathematically convenient circumstance of causal sufficiency, or more broadly, a closed system whereby all of the forces and mechanisms influencing the random variables are either observed, or are ultimately latent variables that are completely determined by observed variables.

\textbf{Closed System Graph Causal Identification Model}: Assume that $\{X(t),Z\}$ are random variables whose interdependencies are fully described y some theoretical DBN defined by a graph $\bar{G}$ and $\tilde f\approx f$, there $\tilde f$ is defined as the transition function given by
\[
p(X_i(t+1)\in A) = f(X(t),X_{j\neq i}(t+1),\{X_i(t-\tau)\}_{\tau=1,...,p},Z)+\epsilon
\]
wherein $\epsilon$ is a zero mean error term. This additive noise model formulation has been leveraged to establish results on the identifiability of the structure $\bar{\mathcal{G}}$~\cite{peters2014causal,hoyer2008nonlinear}. 

The statistical task is as follows: 
\begin{itemize}
    \item Frequentist: Given $\mathcal{S}$, identify the correct ground truth $\bar{\mathcal{G}}$ and a set of parameters that maximizes the likelihood of the data given the model, $\hat{\theta}$. 
    \item Bayesian: Given $\mathcal{S}$ and some background prior uncertainty knowledge over the structure $\pi_G(\bar{\mathcal{G}})$ and parameters $p(\theta\vert \bar{\mathcal{G}})$, find the a posteriori distribution over the graphs $p\left(\bar{\mathcal{G}}\vert \mathcal{S}\right)$ and, hierarchically, the weights $p\left(\theta\vert \bar{\mathcal{G}}, \mathcal{S}\right)$. 
\end{itemize}
As $\left\vert \mathcal{S}\right\vert\to \infty$, it is known that standard scoring and likelihood metrics enable recovery of the ground truth structure and parameters $\left(\bar{\mathcal{G}},\theta^{\bar{\mathcal{G}}}\right)$. However, with the superexponential scaling of possible graph structures 

Finally, let us investigate in more detail the information regarding the DBN dynamics given by~\eqref{eq:likelihood}.

Consider that we have two observed trajectories for three time steps, that is, 
\[
\mathcal{S}=\left\{X^{(1)}(0),X^{(1)}(1),X^{(1)}(2),X^{(1)}(3),X^{(2)}(0),X^{(2)}(1),X^{(2)}(2),X^{(2)}(3)\right\}
\]

We know the trajectories themselves are independent, so we can write the likelihood as a product. The critical consideration now is the treatment of the starting value $X^{(i)}(0)$. It can be taken as an exogenous variable, which would place it in the same role as the conditioned parameters $\theta$ and $\bar{\mathcal{G}}$. Alternatively, a prior of $p\left(X(0)\vert \theta^0, \theta,\bar{\mathcal{G}}\right)$ would specify that a particular DBN is associated with certain starting points. However, notice that we must add an additional parameter $\theta^0$, which would functionally play a similar role as simply conditioning on $X^{(i)}(0)$ itself. So, likelihood can be written to be of the form,
\[
\begin{array}{l}
L\left(\mathcal{S}\vert \theta,\bar{\mathcal{G}}\right)
= p\left(X^{(1)}(1),X^{(1)}(2),X^{(1)}(3)\vert X^{(1)}(0),\theta,\bar{\mathcal{G}}\right)p\left(X^{(2)}(1),X^{(2)}(2),X^{(2)}(3)\vert X^{(2)}(0),\theta,\bar{\mathcal{G}}\right) \\
\qquad = p\left(X^{(1)}(2),X^{(1)}(3)\vert X^{(1)}(1),X^{(1)}(0),\theta,\bar{\mathcal{G}}\right) p\left(X^{(1)}(1)\vert X^{(1)}(0),\theta,\bar{\mathcal{G}}\right)\\ \qquad\qquad \times p\left(X^{(2)}(2),X^{(2)}(3)\vert X^{(2)}(1),X^{(2)}(0),\theta,\bar{\mathcal{G}}\right) p\left(X^{(2)}(1)\vert X^{(2)}(0),\theta,\bar{\mathcal{G}}\right) \\
\qquad = p\left(X^{(1)}(3)\vert X^{(1)}(2),X^{(1)}(0),\theta,\bar{\mathcal{G}}\right) p\left( X^{(1)}(2) \vert X^{(1)}(1),X^{(1)}(0),\theta,\bar{\mathcal{G}}\right) p\left(X^{(1)}(1)\vert X^{(1)}(0),\theta,\bar{\mathcal{G}}\right)\\ \qquad\qquad \times p\left(X^{(2)}(3)\vert X^{(2)}(2),X^{(2)}(0),\theta,\bar{\mathcal{G}}\right) p\left( X^{(2)}(2) \vert X^{(2)}(1),X^{(2)}(0),\theta,\bar{\mathcal{G}}\right) p\left(X^{(2)}(1)\vert X^{(2)}(0),\theta,\bar{\mathcal{G}}\right) \\
\end{array}
\]
However, the latter transitions are independent given the starting point, suggesting that arithmetically we are indeed back to~\eqref{eq:likelihood}. So this is technically correct. 

In order to see why this is still consistent with the intuition that trajectories should have a greater degree of independence, let us continue rewrite the likelihood:
\[
\begin{array}{l}
L\left(\mathcal{S}\vert \theta,\bar{\mathcal{G}}\right)
= p\left(X^{(1)}(3)\vert X^{(1)}(2),X^{(1)}(0),\theta,\bar{\mathcal{G}}\right) p\left( X^{(1)}(2) \vert X^{(1)}(1),X^{(1)}(0),\theta,\bar{\mathcal{G}}\right) p\left(X^{(1)}(1)\vert X^{(1)}(0),\theta,\bar{\mathcal{G}}\right)\\ \qquad\qquad \times p\left(X^{(2)}(3)\vert X^{(2)}(2),X^{(2)}(0),\theta,\bar{\mathcal{G}}\right) p\left( X^{(2)}(2) \vert X^{(2)}(1),X^{(2)}(0),\theta,\bar{\mathcal{G}}\right) p\left(X^{(2)}(1)\vert X^{(2)}(0),\theta,\bar{\mathcal{G}}\right) \\ 
\qquad = p\left(X^{(1)}(3)\vert X^{(1)}(2),\theta,\bar{\mathcal{G}}\right) p\left( X^{(1)}(2) , X^{(1)}(1)\vert X^{(1)}(0),\theta,\bar{\mathcal{G}}\right) \\ \qquad\qquad \times p\left(X^{(2)}(3)\vert X^{(2)}(2),\theta,\bar{\mathcal{G}}\right) p\left( X^{(2)}(2), X^{(2)}(1) \vert X^{(2)}(0),\theta,\bar{\mathcal{G}}\right)
\end{array}
\]
Continuing this through, we can see that as $T\to \infty$, the expression becomes 
\[
p\left(X^{(i)}(T)\vert X^{(i)}(T-1),\theta,\bar{\mathcal{G}}\right) p\left(X^{(i)}(T-1), \cdots ,X^{(i)}(1)\vert X^{(i)}(0),\theta,\bar{\mathcal{G}}\right)
\]
from which we can see the intuition of the circumstance. Asymptotically, the second term approaches the stationary distribution, and the independence assumption becomes valid. Indeed, the structure of the DAG and overall model is expected to be such that the system is long run ergodic. Otherwise, we can consider that for $T$ much longer than the mixing time, this assumption is also valid for most of the transitions. For shorter times of trajectories we can see that:
\begin{enumerate}
    \item The larger the measure of the support, and the more distinct the starting points $X^{(i)}(0)$ are from each other, the longer it can take for the stochastic process to mix to erase the information from initial conditions.
    \item In finite time, the influence of history will depend on the conductance of the Markovian process defined by $(\theta^{\bar{\mathcal{G}}},\bar{\mathcal{G}})$, that is,
    \begin{equation}\label{eq:conduct}
        \phi(\bar{\mathcal{G}}) = \min\limits_{S,S'\subset \bar{\mathcal{G}},\vert S\vert,\vert S' \vert < \vert \bar{\mathcal{G}}\vert /2} 
        \left\{\frac{A(S,S'; \theta,\bar{\mathcal{G}})}{\vert S\vert}
        \right\}
    \end{equation}
    where,
    \[
    A(S,S'; \theta,\bar{\mathcal{G}}) :=\frac{\sum_{i\in S}\sum_{j\in S'} p( X_j(t+1)\vert V_i) }{\vert S\vert }
    \]
    where $V_i$ could be any predecessor in the graph for $X_j(t+1)$. 
\end{enumerate}
This appears in the previous likelihood as follows: we are actually not learning generic trajectories, but those associated with the history of the trajectory, since we are learning conditional distributions. So, in the previous calculation, under the most unfavorable scenario, $(X^{(1)}(1),X^{(1)}(2),X^{(1)}(3))$ and $(X^{(2)}(1),X^{(2)}(2),X^{(2)}(3))$ would correspond to different regions of state space for $X$, that is $X^{(1)}(t)\ge C_1+C_2$ and $X^{(2)}(t) \le C_1 - C_2$, for some large $C_2> 0$, and we learning completely independent transitions that don't inform each other, and moreover, with low spatial correlations, the information gained in the marginal is proportional to $X^{(i)}(0)$.

We can finally conclude from this analysis that the distinction in time transitions and trajectories is only relevant as far as which conditional dependencies they present. Fundamentally, the overall population prior observed for the conditioned variable determines the usefulness of the observations for learning statistical long run averages of the process. 
\subsection{Sample Complexity for Forecasting}
Where the intuition described above arises is in recent results in sample complexity. We shall see that while the arithmetic of~\eqref{eq:likelihood} is still correct for DBNs, there are indeed important distinctions on the sample complexity with respect to the number of different trajectories $N$ and the length of the trajectory $T$ in certain data regimes.

Classically, theoretical analyses of time series sample complexity typically assumed that the trajectory is much longer than the mixing time and by cutting the synthetic burn in period, as such obviates any need to analyze historical dependence. (see the review of the previous results in~\cite{tu2022learning})

We shall report on the theoretical small sample complexity results reported in~\cite{tu2022learning}, which is yet unpublished but extends and otherwise mentions similar recent results in~\cite{xing2022identification,xin2022learning,zheng2020non,dean2020sample}. 

They derive the sample complexity results for learning and identifying a dynamic system,
\begin{equation}\label{eq:ldsforcomplex}
\begin{array}{l}
X(t+1) = A X(t)+B \epsilon(t),\\
Y(t+1) = W X(t)+\xi(t)
\end{array}
\end{equation}
which can be seen a simple Hidden Markov Model and $\epsilon(t),\xi(t)$ are i.i.d. normal random variables. With a goal of fitting a \emph{test} trajectory of length $T'$ (that is, not necessarily equal to $T$), i.e.,
\[
L(\hat{f};T',P_x):=\mathbb{E}_{P_x}\left[\frac{1}{T'}\sum\limits_{t=1}^{T'} \left\|\hat f(X(t))-f_{W}(X(t))\right\|^2\right]
\]
with a minimax risk, i.e., minimizing, algorithmically, the maximal risk associated with the worst case population subsample $P_x\in \mathcal{P}_x$. They compute the guarantees associated with the least squares solution, as defined by the specification of~\eqref{eq:likelihood} to the form given in~\eqref{eq:ldsforcomplex}, with a least squares loss, i.e.,
\[
\hat{W} \in \arg\min\limits_W \sum\limits_{i=1}^N\sum\limits_{t=1}^T \left\| W X^{(i)}(t)-Y^{(i)}(t)\right\|^2
\]
Finally, they require a \emph{trajectory small-ball} assumption, that can be understood as a uniform bound on the covariance matrices associated with the noise in the sequence. 

With this, they present three major results, which are restated here in their informal form.
\begin{theorem}\cite[Theorem 1.1-3]{tu2022learning} 
\begin{enumerate}
   \item If $N\ge n$, $T'\le T$, and the trajectories are drawn from a trajectory small ball distribution, then the excess prediction risk over horizon length $T'$ is $\Theta\left(n/(NT)\right)$
   \item If $N\le n$, $NT\ge n$ and $A$ is marginally unstable and diagonalizable, then the worst case excess prediction risk over horizon length $T'$ is $\Theta\left(n/(NT)\right)\max\left\{nT'/(NT),1\right\}$
   \item If $N\ge n$ and covariate trajectories are such that $A$ is marginally unstable and diagonalizable, then the worst-case excess prediction risk over $T'$ is $\Theta\left(n/(NT) \max\left\{T'/T,1\right\}\right)$
    \end{enumerate}
\end{theorem}
From this Theorem, we can consider that with enough samples, standard rates of sample complexity treating the trajectory length $T$ and the number of trajectories $N$ apply. However, for large relative dimension size of the variable space, the complexity does not scale as well, but is similarly proportional. Finally, when attempting to fit longer trajectories $T'$, we finally see that there is greater benefit towards obtaining data samples with long trajectory lengths over sampling more trajectories.

We report on the one prominent result as far as learning so as to achieve accurate inference on BNs. The classic work~\cite{dasgupta1997sample} reports on a sample complexity, in VC dimension analysis, of modeling a Bayesian Network to be,
\[
\tilde{O}\left(\frac{n^2}{\epsilon^2}\left(n2^k+\log\frac{1}{\delta}\right)\right)
\]
where $\tilde{O}$ suppresses multiplicative terms of $\log(n/\epsilon)$, $\delta$ and $\epsilon$ define the probability of an inference within a small distance of the true outcome, $n$ is the number of variables, and $k$ is the number of potential parents.

\subsection{Sample Complexity for Identification}

The sample complexity given above is for a measure of forecasting error, i.e., excess prediction risk formally. As noted in the Introduction, DBNs are used for a number of purposes. This includes not just forecasting, but also identifying a graph structure that is an interpretative model of potential causal relationships between variables. 

To the best of our knowledge, there are no sample complexity results on graph and causal discovery identification which take separate consideration of trajectories and time steps in the data. Instead we report on a few general recent results on the overall sample complexity for learning a (D)BN as well as a recent result on causal discovery specifically. 

In general, identifying the Bayesian Network is NP-Complete with respect to the number of variables~\cite{chickering1996learning}. It is noted that the number of possible DAGs for 10 variables is greater than \texttt{4$\times$10$^{18}$}~\cite{peters2017elements}.

There are some additional sample complexity results worth reporting from the literature.

The work~\cite{ordyniak2013parameterized} presents poly-time identifiability in the case of bounded treewidth or acyclic super-structure, and otherwise confirms NP-Hardness of search with respect to data. A creative recent work~\cite{geduk2022practical} uses models from physiology to argue for $O\left(M^k\right)$ practical complexity, with $M$ the cardinality of a discrete valued network and $k$ is the number of potential parents. 

More favorable results are presented for linear SEMs with a recent algorithm that improves the sample complexity to $O\left(n^2\log k\right)$ in the case of sub-Gaussian errors and $O\left(n^2k^{2/m}\right)$ for $4m$-bounded moment errors. 

Finally, the work~\cite{wadhwa2021sample} considers sample complexity of causal discovery specifically, which runs at the number of samples required being $O\left(n! l^{3n/8}\right)$, where $l$ is the cardinality of the possible random variable values in a discrete network. 

\subsection{Formalization: Open System Forecasting Model}

In practice, in many cases wherein DBNs are employed for modeling, understanding and forecasting, the underlying system is not completely closed, as in a physics experiment, or deliberately marginalized, as in a randomized clinical trial. Instead, it models a complex and often infinite dimensional system, with intricate and impossible-to-know interactions with the environment. With the presence of unknown confounders, causal sufficiency isn't satisfied. Moreover, it can happen that multiple structures and parameters become equally effective at accurately modeling the process, even highly distinct ones suggesting distinct causal mechanisms.

For instance, DBNs are often used for predictive maintenance, as in~\cite{amin2019fault,yu2013novel}. By an appropriate representation of the underlying complex engineering system as distilled into some low dimensional latent structure, one can develop DBNs to monitor signals of deterioration or damage in the system as based on the historical transitions over time in performance.

An interesting formalism of this is given in~\cite{esmaeil2015dynamic}. For some underlying stochastic process $\dot{Y}=g(Y(t),W(t))$ with (e.g., Brownian) noise $W(t)$ where $Y\in\mathcal{Y}$ is very high, if not infinite, dimensional, one can consider a DBN model as a finite dimensional reduced order model of the system, and one that maximizes the information relevance towards maintenance. Formal guarantees are provided as far as probabilistic invariance, that is,
\[
p\left(X(t)\in A,\forall \, t\in[T]\right) \approx p\left(Y(t)\in \tilde{A},\forall 0\le t\le T\right)
\]
indicating the potential for DBNs to serve as useful indicators of higher level properties of stochastic processes, regardless of the fundamental impossibility of formal causal structure identification in such cases. These considerations, and the study of DBNs without causal sufficiency in general, is broadly limited, however.

\section{Learning, Loss Criteria and Constraint Definitions} 
Now we will proceed to present some of the analytical expressions associated with learning DBNs. Recall that we assume we have a sample of $N$ trajectories over time horizon $T$, that is, we restate~\eqref{eq:sample},
\[
\mathcal{S}=\cup_{n=1}^N \mathcal{T}^n=\{ Z^{(n)},X^{(n)}(0),X^{(n)}(1), X^{(n)}(2), ..., X^{(n)}(T)\}_{n=1,...,N}
\]

\subsection{Criteria}
In order to ascertain the goodness of fit of different DBN graph structures, a score function serves as an objective in an optimization process. The score function is meant to evaluate the statistical accuracy of a model together with its parsimony. In performing structure learning as guided by a score, we are performing a likelihood, or some maximum a posteriori maximization, in the process of traversing the decision landscape of structures. 

Selection criteria for models appears in both the BN/PGM as well as the time series modeling literature. In~\cite{mcquarrie1998regression} a thorough exploration of the evaluation and computation of various criteria is presented for a range of different time series models. 
In~\cite{douc2014nonlinear} it is recommended to use a general form for an information criterion to evaluate possible networks is, for $N$ samples (not distinguishing between trajectories and time points) and a total $k$ parameters:
\begin{equation}\label{eq:criterion}
\Delta_{k,N} = -2\log L_k+C_{k,N}
\end{equation}
where $L_k$ is the likelihood of the data given the model and parameters and $C_{k,N}$ is a parsimony term with the following forms:
\begin{itemize}
    \item AIC  $C_{k,N}=2k$
    \item AICc $C_{k,N}=\frac{N+k}{N-k-2}$
    \item BIC $C_{k,N}=k\log N$
\end{itemize}

From an alternative perspective, we consider how the Likelihood can be computed and maximized. Writing the quantity as $L_k(\{X\}\vert \Theta,\mathcal{G})$ we see that it depends on both the structure $\mathcal{G}$ and the parameters $\Theta$. Observing that 1) the parameter space itself depends on the graph structure, and 2) the likelihood is differentiable with respect to the parameters whereas the structure is a discrete quantity that presents, fundamentally, a combinatorial optimization problem. 

A natural approach, which can be described as a sort of hierarchical frequentist method, would be to maximize $L_k(\{X\}\vert \Theta(\mathcal{G}),\mathcal{G})$ with respect to $\mathcal{G}$. In doing so, however, to evaluate the likelihood given $\mathcal{G}$, one must compute $\Theta(\mathcal{G})$ that is,
\[
\Theta(\mathcal{G}) = \arg\max_{\theta} L_k(\{X\}\vert \theta,\mathcal{G})
\]
where conditioning on $\mathcal{G}$ enforces the appropriate structure of $\theta$ to be zero. Computationally, one can also pursue a \emph{one shot} approach wherein the structure and weights are maximized simultaneously, with some appropriate functional enforcement of consistency of the structure with the zero sparsity structure of weights. While formally, a global maximum obtained with a hierarchical procedure is going to be the same equivalent pair $(\mathcal{G}^*,\Theta^*)$, in practice the choice does carry a distinction as far as tendencies and patterns of the solutions of the optimization procedure. 

From a frequentist perspective, the term $C_{k,N}$ is a regularizer chosen instrumentally to prioritize sparser solutions. Sparse basis recovery has long been an important area of statistical research (e.g.~\cite{wainwright2009information}, with a popular monograph (with a chapter on PGMs) \cite{wainwright2019high} (see also \cite{wainwright2008graphical}). 

In taking a Bayesian perspective, one can consider a functional / computational equivalence by considering $C_{k,N}$ as the prior, with varying strength corresponding to the degree of confidence in the prior. 
Formally, the score function is the marginal posterior of the candidate structure given the data, that is $p(\mathcal{G}\vert \{X\})$. This, of course, is unknown in form but can be appropriately decomposed with the parameters serving as ``nuisance'' vectors, i.e., 
\[
p(\mathcal{G}\vert \{X\}) = \frac{\int_{\theta} p(\{X\} \vert \theta,\mathcal{G}) p (\theta \vert \mathcal{G}) p(\mathcal{G}) d\theta }{\int_{\mathcal{G}}\int_{\theta} p(\{X\} \vert \theta,\mathcal{G}) p (\theta \vert \mathcal{G}) p(\mathcal{G}) d\theta d\mathcal{G}}
\]
For particular kinds of parametrized DBNs, computing this posterior can be done in closed form without evaluating explicitly the terms involving $\theta$. For a classic discussion on the statistical intuition, motivation, and some formulations of Bayesian criteria, see~\cite{heckerman1995learning}.

\subsection{Likelihood Calculations}
A common assumption made in the literature ~\cite{koller2009probabilistic,geiger2002parameter} is that of global parameter independence. That is, it holds that the parameters $(\theta\vert \bar{\mathcal{G}})$ can be decomposed to be separable across the transitions for each variable $X_i$, i.e. using the notation $\theta^{\bar{\mathcal{G}},i}$ to indicate parameters associated with the transition step for variable $X_i(t+1)$,
\begin{assumption}\label{as:paramind}
    It holds that,
    \[
    p\left(\theta^{\bar{\mathcal{G}}} \vert \bar{\mathcal{G}}\right)
     = \prod\limits_{i\in[n]} p\left(\theta^{\bar{\mathcal{G}},i} \vert \bar{\mathcal{G}}\right)
    \]
    and that, for any data sample $\mathcal{S}$,
    \[
    p\left(\mathcal{S}\vert \theta_G,\bar{\mathcal{G}} \right) = 
    \prod\limits_{i\in[n]} p\left(\mathcal{S}\vert \theta^{\bar{\mathcal{G}},i}, \bar{\mathcal{G}}\right)
    \]
\end{assumption}

This assumption significantly eases computation, as the solution can be obtained by maximizing the likelihood for the set of parameters separately for each $i$. Now, to obtain the expression for the total likelihood in this case, perform the usual condition dependence chain 
\[
\begin{array}{l}
p(X(t+1),X(t),...,X(0)\vert \theta) = p(X(t+1)\vert X(t),X(t-1),..., X(1)\vert \theta) \\ \qquad =p(X(t+1)\vert X(t),\theta)p(X(t)\vert X(t-1),\theta),...,p(X(0))
\end{array}
\]
This yields the complete expression below, where we now include all the structured dependencies.
\begin{equation}
\begin{array}{l}
p\left(\mathcal{S}\vert \theta^{\bar{\mathcal{G}},i},\bar{\mathcal{G}} \right) = \prod\limits_{n=1}^N \prod\limits_{s=1}^{T}  p\left(X^{(n)}_i(T-s+1)\vert Z^{(n)},\{X_j^{(n)}(T-s)\}_{j\in dpa_d(i)},\right.\\\qquad\qquad\left.\{X^{(n)}_j(T-s+1)\}_{j \in dpa_s(i)},\{X_i^{(n)}(T-s-\tau)\}_{\tau\in dpa_{\tau}(i)}, \theta^{\bar{\mathcal{G}},i},\bar{\mathcal{G}}\right) 
\end{array}
\end{equation}
We will also use $\bar{\Xi}_i$ to indicate the variables corresponding to the set of dependencies $dpa(i)$
\subsubsection{Binary Variables}
In the case wherein all variables $\{X_i,Z\}$ are valued $\{0,1\}$ sampled from a Bernoulli distribution, this presents the simplest calculation, recalling the definition of the transition model.

More significantly, here, we shall see that the more complex representation permits for closed form computation of the marginal posterior of the structure. 

To begin with, the simple linear model~\eqref{eq:binarylinear}.
In this case we write,
\[
p(X_i(t+1)=1;\theta^{\bar{\mathcal{G}},i}) 
 = \sigma\left(\theta^{\bar{\mathcal{G}},i}_0+\sum\limits_{j\in dpa(i)} \theta^{\bar{\mathcal{G}},i}_{j} V_j\right)
\]
From this functional form we can obtain, recalling generically $V_{dpa(i)}$ for any parents, by any of the dependencies, of the variable $i$. 
\begin{equation}\label{eq:binarylikelihood}
\begin{array}{l}
p\left(\mathcal{S}\vert \theta^{\bar{\mathcal{G}},i},\bar{\mathcal{G}} \right) =
\prod\limits_{n=1}^N \prod\limits_{t=0}^{T-1} \left[
\mathbf{1}(X^{(n)}_i(t+1)=1)P(X^{(n)}_i(t+1)=1\vert V^{(n)}_{dpa(i)} ;\theta)
\right. \\ \qquad\qquad \left.+\mathbf{1}(X^{(n)}_i(t+1)=0)P(X^{(n)}_i(t+1)=0\vert V^{(n)}_{dpa(i)} ;\theta)\right]
\end{array}
\end{equation}
Now we take a logarithm of the expression, turning the products into sums,
\begin{equation}\label{eq:binaryloglikelihood}
\begin{array}{l}
\log \left(p\left(\mathcal{S}\vert \theta^{\bar{\mathcal{G}},i},\bar{\mathcal{G}} \right)\right) =
\sum\limits_{n=1}^N \sum\limits_{t=0}^{T-1} \left[
\mathbf{1}(X^{(n)}_i(t+1)=1)\left[\theta^{\bar{\mathcal{G}},i}_0+\sum\limits_{j\in dpa(i)} \theta^{\bar{\mathcal{G}},i}_{j} V_j\right.
\right. \\ 
\qquad\qquad \left.-\log\left(1+\exp\left\{\theta^{\bar{\mathcal{G}},i}_0+\sum\limits_{j\in dpa(i)} \theta^{\bar{\mathcal{G}},i}_{j} V_j\right\}\right) \right]\\
\qquad\qquad \left.\left. 
-\mathbf{1}(X^{(n)}_i(t+1)=0)\log\left(1+\exp\left\{\theta^{\bar{\mathcal{G}},i}_0+\sum\limits_{j\in dpa(i)} \theta^{\bar{\mathcal{G}},i}_{j} V_j\right\}\right)\right]\right]
\end{array}
\end{equation}
In this case, the maximum likelihood cannot be computed in closed form, and numerical methods must be used. The similar situation holds for computing a Bayesian score under this model restiction.

Now we consider the full combinatorial representation as defined by ~\eqref{eq:genericthetaxitwo}, which, with binary outcomes, simplifies to:
\begin{equation}\label{eq:genericthetaxitwo}
\begin{array}{l}
p(X_i(t+1)=1 \vert V_{dpa(i)},\theta_i) := \theta_i^{\xi(V_{dpa(i)})},\\
\xi\in \Xi_i,\,
\Xi_i:=\Xi^d_i\times \Xi^s_i:= \prod_{j\in dpa_t(i)} \mathbb{Z}_{2}^+ \times \prod_{j\in dpa_z(i)} \mathbb{Z}_2^+
\end{array}
\end{equation}
Indeed this corresponds to the local parameter independence and the unrestricted multinomial conditions that facilitates the closed form computation for the Bayesian Dirichlet scores. To this end,
we now extend the presentation in~\cite{heckerman1995learning} (see also~\cite{heckerman2008tutorial}) to include the contribution of the static $Z$ variables to the model. 

First we begin by writing the full expression for the likelihood and computing the likelihood-maximizing parameter values, making use of the modeling representation in~\eqref{eq:genericthetaxitwo}.

We introduce one more piece of notation, indicating the set of dynamic variables that contribute in the DAG structure to node $i$, 
\[
V^{(n)}_{i,d}(t) = \{X_j^{(n)}(t-1)\}_{j\in dpa_d(i)} \cup \{X_j^{(n)}(t)\}_{j\in dpa_s(i)} \cup \{X_i^{(n)}(t-\tau)\}_{\tau\in dpa_{\tau}(i)}
\]
this distinguishes the dynamic variables from the static ones.
\begin{equation}\label{eq:binlikderivation}
\begin{array}{l}
p\left(\mathcal{S}\vert \theta,\bar{\mathcal{G}})\right)=
\prod\limits_{i\in[n_x]}
\prod\limits_{n=1}^N \prod\limits_{t=0}^{T-1} \left[
p(X_i(t+1)=1 \vert V_{dpa(i)},\theta_i) X^{(n)}_i(t+1)\right. \\ \qquad\qquad \qquad\qquad  \left.+(1-p(X_i(t+1)=1 \vert V_{dpa(i)},\theta_i))(1-X^{(n)}_i(t+1)
\right] \\ 
= \quad \prod\limits_{i\in[n_x]} \prod\limits_{\xi^d \in \Xi^d_i}
\prod\limits_{\xi^s \in \Xi^s_i}
\prod\limits_{n=1}^N
\prod\limits_{t=0}^{T-1} \left[
p(X_i(t+1)=1 \vert \xi,\theta_i) X^{(n)}_i(t+1) \mathbf{1}(\xi^d = V^{(n)}_{i,d}(t+1))\mathbf{1}(\xi^s = Z^{(n)}) \right. \\
\qquad\qquad \left.+(1-p(X_i(t+1)=1 \vert V_{dpa(i)},\theta_i))(1-X^{(n)}_i(t+1))  \mathbf{1}(\xi^d = V^{(n)}_{i,d}(t+1))\mathbf{1}(\xi^s = Z^{(n)}) 
\right] \\
= \quad \prod\limits_{i\in[n_x]} \prod\limits_{\xi^d \in \Xi^d_i}
\prod\limits_{\xi^s \in \Xi^s_i}
\prod\limits_{n=1}^N
\prod\limits_{t=0}^{T-1} \left[
\theta_i^{\xi(V_{dpa(i)})} X^{(n)}_i(t+1) \mathbf{1}\left(\xi^d = V^{(n)}_{i,d}(t+1)\right)\mathbf{1}\left(\xi^s = Z^{(n)}\right) \right. \\
\qquad\qquad \left.+\left(1-\theta_i^{\xi(V_{dpa(i)})}\right)
\left(1-X^{(n)}_i(t+1)\right)  \mathbf{1}\left(\xi^d = V^{(n)}_{i,d}(t+1)\right)\mathbf{1}\left(\xi^s = Z^{(n)}\right) 
\right] 
\end{array}
\end{equation}

Let $\mathbf{N}(A; \mathcal{C})$ be the counting operator of the number of elements of $\mathcal{C}$ that satisfy the condition given by $A$.
Now take the logarithm of the likelihood expression and obtain a sum-separable set of terms for the log likelihood of each parameter, and perform generative learning to find the parameters. Specifically,
\[
\begin{array}{l}
\log p\left(\mathcal{S}\vert \theta_i^{\xi(V_{dpa(i)})},\bar{\mathcal{G}}\right)  \\ \quad = \mathbf{N}\left(\left[(X^{(n)}_i(t+1)=1)\cap \left(V^{(n)}_{i,d}(t+1)\times Z^{(n)}=\xi(V_{dpa(i)})\right)\right];\mathcal{S}\right)\log\left(\theta_i^{\xi(V_{dpa(i)})}\right)\\ 
\qquad \qquad
+\mathbf{N}\left(\left[(X^{(n)}_i(t+1)=0)\cap (V^{(n)}_{i,d}(t+1)\times Z^{(n)}=\xi(V_{dpa(i)})\right];\mathcal{S}\right)
\log\left(1-\theta_i^{\xi(V_{dpa(i)})}\right)
\end{array}
\]
From which the natural maximum likelihood estimate can be formed:
\begin{equation}\label{eq:maxlikbin}
    \hat{\theta}_i^{\xi(V_{dpa(i)})} = \frac{\mathbf{N}\left(\left[(X^{(n)}_i(t+1)=1)\cap \left(V^{(n)}_{i,d}(t+1)\times Z^{(n)}=\xi(V_{dpa(i)})\right)\right];\mathcal{S}\right)}{\mathbf{N}\left(\left[\left(V^{(n)}_{i,d}(t+1)\times Z^{(n)}=\xi(V_{dpa(i)})\right)\right];\mathcal{S}\right)}
\end{equation}
Note that in this case, the counts are over both the samples of trajectories and the time points between them. Observe the role of the static variables $Z$ as simply interacting covariates in the form. Thus, when $Z$ is of a mechanistic form that mediates the transitions, its influence is absorbed as simply an added dimension to the parameter space. We can, however, force a distinction between dynamic and static effects if we assume their causal independence. This would correspond to a kernel transition of the form:
\begin{equation}\label{eq:genericthetaxitwoinddystat}
\begin{array}{l}
p\left(X_i(t+1)=1 \vert V_{dpa(i)},\theta_i\right) := \theta_i^{\xi^d(V_{dpa_t(i)})} \theta_i^{\xi^s(Z_{dpa_z(i)})} 
\end{array}
\end{equation}
where $V_{dpa_t(i)}$ denotes the full set of time-dependent variables that influence $i$. 
It can be seen that we can obtain the maximum likelihood estimates as,
\begin{equation}\label{eq:maxlikbindystat}
\begin{array}{l}
    \hat{\theta}_i^{\xi^d(V_{dpa(i)})} = \frac{\mathbf{N}\left(\left[(X^{(n)}_i(t+1)=1)\cap \left(V^{(n)}_{i,d}(t+1)=\xi(V_{dpa_t(i)})\right)\right];\mathcal{S}\right)}{\mathbf{N}\left(T\left[\left(V^{(n)}_{i,d}(t+1)=\xi(V_{dpa_t(i)})\right)\right];\mathcal{S}\right)} \\ 
    \hat{\theta}_i^{\xi^s(Z_{dpa_z(i)})} = \frac{\mathbf{N}\left(\left[(X^{(n)}_i(t+1)=1)\cap \left( Z^{(n)}=\xi(Z_{dpa_z(i)})\right)\right];\mathcal{S}\right)}{\mathbf{N}\left(\left[\left(Z^{(n)}=\xi(Z_{dpa_z(i)})\right)\right];\mathcal{S}\right)} = 
    \frac{\mathbf{N}\left(\left[(X^{(n)}_i(t+1)=1)\cap \left( Z^{(n)}=\xi(Z_{dpa_z(i)})\right)\right];\mathcal{S}\right)}{T\mathbf{N}\left(\left[\left(Z^{(n)}=\xi(Z_{dpa_z(i)})\right)\right];n\in [N]\right)}
\end{array}
\end{equation}
From this we can see indeed that with independent causal influence, the estimate for the parameters governing the static nodes $Z$'s influence carries more statistical power, with an effective sample size scaled by $T$. 

Now we present the computation of the Bayesian Dirichlet scores.
This amounts to computing the marginal posterior of the structure by performing an integration treating parameter as nuisance. This is derived, for instance, in~\cite{geiger2002parameter}, and used in the popular integer BN structure learner GOBNILP~\cite{cussens2020gobnilp}. The marginal posterior of the structure is given by:
\[
p\left(\bar{\mathcal{G}} \vert \mathcal{S}\right) =
\int_{\theta} p(\mathcal{S}\vert \theta^{\bar{\mathcal{G}}},\bar{\mathcal{G}})p(\theta^{\bar{\mathcal{G}}}\vert \bar{\mathcal{G}}) d\theta^{\bar{\mathcal{G}}}
\]
In order to compute the BDe, we need a prior on the weights, which we write as a Dirichlet distribution,
\[
p(\theta^{\bar{\mathcal{G}}}\vert {\bar{\mathcal{G}}}) = 
\prod_{i\in[n]}\prod_{\xi\in \Xi_i} \frac{\Gamma\left(\alpha^{i,\xi}_0+\alpha^{i,\xi}_1\right)}{\Gamma\left(\alpha^{i,\xi}_0\right)+\Gamma\left(\alpha^{i,\xi}_1\right)}\theta^{\alpha^{i,\xi}_0}_{i,\xi,0}
\theta^{\alpha^{i,\xi}_1}_{i,\xi,1}
\]
Recalling the expression for~\eqref{eq:binlikderivation}, we can see that the BDe can be computed by,
\[
\begin{array}{l}
p\left(\bar{\mathcal{G}} \vert \mathcal{S}\right) \\ \quad = \prod\limits_{i\in[n_x]} 
\prod\limits_{\xi^d \in \Xi^d_i}
\prod\limits_{\xi^s \in \Xi^s_i}
\prod\limits_{n=1}^N
\prod\limits_{t=0}^{T-1} \int_{\theta} \left[
\theta_i^{\xi(V_{dpa(i)})} X^{(n)}_i(t+1) \mathbf{1}\left(\xi^d = V^{(n)}_{i,d}(t+1)\right)\mathbf{1}\left(\xi^s = Z^{(n)}\right) \right. \\
\qquad\qquad \left.+\left(1-\theta_i^{\xi(V_{dpa(i)})}\right)
\left(1-X^{(n)}_i(t+1)\right)  \mathbf{1}\left(\xi^d = V^{(n)}_{i,d}(t+1)\right)\mathbf{1}\left(\xi^s = Z^{(n)}\right) 
\right] \\ \qquad \qquad \qquad \times \frac{\Gamma\left(\alpha^{i,\xi}_0+\alpha^{i,\xi}_1\right)}{\Gamma\left(\alpha^{i,\xi}_0\right)+\Gamma\left(\alpha^{i,\xi}_1\right)}\theta_{i,\xi}^{\alpha^{i,\xi}_0}
\theta_{i,\xi}^{\alpha^{i,\xi}_1} d\theta \\
= \prod\limits_{i\in[n_x]} 
\prod\limits_{\xi^d \in \Xi^d_i}
\prod\limits_{\xi^s \in \Xi^s_i} \frac{\Gamma\left(\alpha^{i,\xi}_0+\alpha^{i,\xi}_1\right)}{\Gamma\left(\alpha^{i,\xi}_0\right)+\Gamma\left(\alpha^{i,\xi}_1\right)}  \times 
\int_{\theta} \theta_{i,\xi,0}^{\alpha^{i,\xi}_0+\mathbf{N}_{i,\xi}
-\mathbf{N}_{i,\xi,1}}
\theta_{i,\xi,1}^{\alpha^{i,\xi}_1+\mathbf{N}_{i,\xi,1}} d\theta \\
= \prod\limits_{i\in[n_x]} 
\prod\limits_{\xi^d \in \Xi^d_i}
\prod\limits_{\xi^s \in \Xi^s_i} \frac{\Gamma\left(\alpha^{i,\xi}_0+\alpha^{i,\xi}_1\right)}{\Gamma\left(\alpha^{i,\xi}_0\right)+\Gamma\left(\alpha^{i,\xi}_1\right)}  \times 
\frac{(\alpha^{i,\xi}_1+\mathbf{N}_{i,\xi,1})}{(\mathbf{N}_{i,\xi}+\alpha^{i,\xi}_1+\alpha^{i,\xi}_0)}
\end{array}
\]
where 
\[
\begin{array}{l}
\mathbf{N}_{i,\xi,1} =\mathbf{N}\left(\left[(X^{(n)}_i(t+1)=1)\cap \left(V^{(n)}_{i,d}(t+1)\times Z^{(n)}=\xi(V_{dpa(i)})\right)\right];\mathcal{S}\right),\\
\mathbf{N}_{i,\xi} =\mathbf{N}\left(\left[V^{(n)}_{i,d}(t+1)\times Z^{(n)}=\xi(V_{dpa(i)})\right];\mathcal{S}\right)
\end{array}
\]
and with dynamic-static causal influence independence, the score becomes,
\begin{equation}\label{eq:bde}
\begin{array}{l}
p\left(\bar{\mathcal{G}} \vert \mathcal{S}\right) = 
\prod\limits_{i\in[n_x]} 
\prod\limits_{\xi^d \in \Xi^d_i}
\frac{\Gamma\left(\alpha^{i,\xi^d}_0+\alpha^{i,\xi^d}_1\right)}{\Gamma\left(\alpha^{i,\xi^d}_0\right)+\Gamma\left(\alpha^{i,\xi^d}_1\right)}  \times 
\frac{(\alpha^{i,\xi^d}_1+\mathbf{N}_{i,\xi^d,1})}{(\mathbf{N}_{i,\xi^d}+\alpha^{i,\xi^d}_1+\alpha^{i,\xi^d}_0)} \\ \qquad \times 
\prod\limits_{\xi^s \in \Xi^s_i} \frac{\Gamma\left(\alpha^{i,\xi^s}_0+\alpha^{i,\xi^s}_1\right)}{\Gamma\left(\alpha^{i,\xi^s}_0\right)+\Gamma\left(\alpha^{i,\xi^s}_1\right)}  \times 
\frac{(\alpha^{i,\xi^s}_1+\mathbf{N}_{i,\xi^s,1})}{(\mathbf{N}_{i,\xi^s}+\alpha^{i,\xi^s}_1+\alpha^{i,\xi^s}_0)}
\end{array}
\end{equation}
Thus, for the DBN case, computing the above amounts to evaluating the BD score. We observe, in addition, that this derivation indicates how one can sample from the posterior distribution of the weights given the structure that a learner identifies as maximizing the desired score. Indeed the posterior of the weights given the structure is shown above, it is the expression under the integral sign, i.e.,
\begin{equation}\label{eq:dirweightpost}
p\left(\theta \vert \bar{\mathcal{G}},\mathcal{S}\right) =
\prod\limits_{i\in[n_x]} 
\prod\limits_{\xi^d \in \Xi^d_i}
\prod\limits_{\xi^s \in \Xi^s_i} \frac{\Gamma\left(\alpha^{i,\xi}_0+\alpha^{i,\xi}_1\right)}{\Gamma\left(\alpha^{i,\xi}_0\right)+\Gamma\left(\alpha^{i,\xi}_1\right)}  \theta_{i,\xi,0}^{\alpha^{i,\xi}_0+\mathbf{N}_{i,\xi}
-\mathbf{N}_{i,\xi,1}}
\theta_{i,\xi,1}^{\alpha^{i,\xi}_1+\mathbf{N}_{i,\xi,1}} 
\end{equation}

\subsubsection{Gaussian DBNs}

Now we present the derivation of the likelihood and Bayesian criterion (BGe) for DBNs with Gaussian models. The development follows~\cite{geiger2002parameter} and extends their derivation in two ways. First we perform the recursion for computing the entire trajectory time data. Second, we include a specific parametrization and show how one can simultaneously perform the recursion to obtain a posterior of the weights. We, however, simplify our model to only include Markovian influence, and not lagged autoregressive effects.

We apply the model in~\cite{geiger2002parameter} to~\eqref{eq:gausslinear} to obtain the following transition likelihood function for the first step and prior for both the overall likelihood transition and the parameters themselves:
\begin{equation}\label{eq:gausslinearb}
\begin{array}{l}
p(X_i(1)\cup X_{j\in dpa_d}(0)\cup X_{j\in dpa_s}(1)\cup Z_{j\in dpa_z}\vert \beta, dpa_d(i)\cup dpa_s(i)\cup dpa_{z}(i)) = \mathcal{N}\left(\mu(0), W\right)\\
\mu(0) = \begin{pmatrix} \mu^x_i(1;0) &  \mu^x_{j\in dpa_d(i)}(0) & \mu^x_{j\in dpa_s(i)}(0) & \mu^z_{j\in dpa_z} (0) \end{pmatrix}^T \\
\mu^x(0)\in\mathbb{R}^{n_x},\,\mu^z(0)\in\mathbb{R}^{n_z} \\
\mu^x_i(1;0) \sim \beta^0+\sum\limits_{j\in dpa_d(i)} \beta^d_{i,j} X_j(0)+\sum\limits_{j\in dpa_s(i)} \beta^s_{i,j} X_j(1) + \sum\limits_{j\in dpa_z(i)} \beta^z_{i,j} Z_j \\
(\beta^0,\beta^d,\beta^s,\beta^z)\sim\mathcal{N}(\eta(0),\psi\Upsilon(0))
\end{array}
\end{equation}
Now consider that the variables have corresponding priors marginal:
\begin{equation}\label{eq:gaussinitial}    X(0),Z \sim \mathcal{N}\left((\mu_x(0),\mu_z(0)),\{\Sigma(0),\Sigma_z\}\right)
\end{equation}
this will be also used to derive the corresponding equivalent posterior analysis. Note that the DAG structure is important for the sensibility of these definitions. 

Let us define $W$. The parameter prior introduces a normal-Wishart distribution on the mean with precision matrix $T$, dropping the $i$ dependence
\begin{equation}\label{eq:priorbge}
    \begin{array}{l}
p\left(\mu(0)\vert W, \bar{\mathcal{G}}\right) = \mathcal{N}\left(\nu(0),\alpha_{\mu} W\right) \\
\nu(0) = \begin{pmatrix} \mu^+(0):= \eta_0 + \eta_d\cdot \mu^x_{j\in dpa_d}(0)  + \eta_s\cdot \mu^x_{j\in dpa_s} +\eta_z\cdot\mu^z_{j\in dpa_z(i)}(0) \\ \mu^x_{j\in dpa_d}(0) \\ \mu^x_{j\in dpa_s}(0) \\ \mu^z_{j\in dpa_z(i)}(0) \end{pmatrix} \\ 
p\left(W \vert \bar{\mathcal{G}}\right) = c(n_i(1),\alpha_w)\vert T \vert^{\alpha_w/2}\vert W \vert^{(a_{\alpha}-n_i(1)-1)/2} e^{-1/2 \mathop{tr}(TW)}\equiv \text{Wishart}\left( W\vert \alpha_{w},T\right)\\
c(n_i(1),\alpha_w):=\left(2^{\alpha_w n/2} \pi^{n(n-1)/4}\prod\limits_{i=1}^{n_x} \Gamma\left(\frac{\alpha_w+1-i}{2}\right)\right)^{-1} \\
n(1) = 1+\vert dpa(i)\vert \\
\alpha_{\mu} W =\\ \psi \begin{pmatrix}  U(0) & U_d(0) & .. \\ 
.. & .. & .. \\ 
.. & .. & U_z(0)\end{pmatrix} \\
U(0) = (\Upsilon(0)\, /\, \Upsilon_{\setminus 0}(0))(\Upsilon(0)\, /\, \Upsilon_{\setminus 0}(0))^T\\
U_d(0) = \mu_{dpa_d(i)}^x(0)(\Upsilon(0)\, /\, \Upsilon_{\setminus 0}(0))(\Upsilon(0)\, /\, \Upsilon_{\setminus d}(0))^T \\ 
U_z(0) = \mu_{dpa_z(i)}^z(0)(\Upsilon(0)\, /\, \Upsilon_{\setminus 0}(0))(\Upsilon(0)\, /\, \Upsilon_{\setminus d}(0)) \\
...
    \end{array}
\end{equation}
where $A/B$ denotes the Schur complement of $A$ with respect to $B$.
In~\cite[Theorem 4 and Theorem 5]{geiger2002parameter} it is shown that parameter independence is preserved through the computation of the posterior. Note that the posterior now is with respect to all the data that is present in a transition. The DAG structure ensures that $\nu(0)$ is well defined as a vector, rather than implicitly as a function of $\mu^x_{j\in dpa_s(i)}(0)$. Finally, the last line related the two models together, indicating how the Wishart distribution arises from the parameter distribution, in this case.

With this, we obtain the joint likelihood expression:
\begin{equation}\label{eq:gausslike}
\begin{array}{l}
    p\left(\mathcal{S}\vert \beta^{\bar{\mathcal{G}}},\bar{\mathcal{G}}\right) = 
     
\prod\limits_{n=1}^N
\prod\limits_{t=0}^{T-1} \prod\limits_{i\in[n_x]} p\left(X^{(n)}_i(t+1)\cup V^{(n)}_{j\in dpa(i)}(t+1)\vert \theta^{\bar{\mathcal{G}}},\bar{\mathcal{G}} \right) \\ 
=  \prod\limits_{i\in[n_x]}\prod\limits_{t=0}^{T-1} 
\prod\limits_{n=1}^N
 p\left(X^{(n)}_i(t+1)\cup  \{X^{(n)}_j(t)\}_{j\in dpa_d(i)},\{X_j(t+1)\}_{j\in dpa(i)},\{Z_j\}_{j\in dpa_z(i)}\vert \mu_x(0),\mu_z(0),\eta(0),\bar{\mathcal{G}} \right) \\
 := \prod\limits_{i\in[n_x]}\prod\limits_{t=0}^{T-1} 
\prod\limits_{n=1}^N
 p\left(\mathcal{S}^{(n)}_i(1)\vert \mu_x(0),\mu_z(0),\eta(0),\bar{\mathcal{G}} \right)
\end{array}
\end{equation}
for mean $\mu$ and nowhere singular covariance matrix $W$. 

Now, with this redundant embedding in both prior in the variable space and parameter space, we deduce how to compute the posterior of the distribution distribution of the data $\mu$ and $W$ from~\cite{geiger2002parameter} for $T=2$, and subsequently, compute the posterior of the parameters in the model, while showing it is equivalent by a straightforward Bayesian posterior propagation. After deriving the base case $T=2$, we continue with the induction for $T$ to $T+1$, in order to derive the final posterior of the data, from which we can compute the marginal likelihood of the structure, as well as sample the final posterior values. 

Now from the original we know that the likelihood of the data can be given by:
\begin{equation}\label{eq:posteriordata}
\begin{array}{l}
    p\left(\mu(1) \vert W,\mathcal{S}^{(n)}_i(1),\bar{\mathcal{G}}\right) \sim 
    \mathcal{N}(\nu(1),(\alpha_w+N)W(1)),\\
    W(1)
    \sim \text{Wishart}(\alpha_w+N,R(1)) \\
    \nu(1) = \begin{pmatrix} \mu^+(1) \\ \mu^x_{j\in dpa_d(i)}(1) \\ \mu^x_{j\in dpa_s(i)}(1) \\ \mu^z_{j\in dpa_z(i)}(1) \end{pmatrix} := \frac{1}{\alpha_{\mu}+N} \left[\alpha_{\mu}\nu(0)+N\bar{\nu}(1)\right]\\
    R = T + S_N(1)+\frac{\alpha_{\mu}N}{\alpha_{\mu}+N}(\nu(0)-\bar{\nu}(1))(\nu(0)-\bar{\nu}(1))^T \\
    \bar{\nu}(1) = \begin{pmatrix} \bar{\nu}(1;0) \\ \bar{\nu}^x_{j\in dpa_d(i)}(0) \\ \bar{\nu}^x_{j\in dpa_s(i)}(0) \\ \bar{\nu}^z_{j\in dpa_z(i)}\end{pmatrix} := \begin{pmatrix} \frac{1}{N}\sum\limits_{n=1}^N X^{(n)}_i(1) \\ 
    \frac{1}{N}\sum\limits_{n=1}^N X^{(n)}_{j\in dpa_d}(0) \\ \frac{1}{N}\sum\limits_{n=1}^N X^{(n)}_{j\in dpa_s(i)}(1) \\ \frac{1}{N}\sum\limits_{n=1}^N Z^{(n)}_{j\in dpa_z(i)} \end{pmatrix} \\
\end{array}
\end{equation}

Now, before we proceed with the next time step, let us define the propagation in the hyperparameters, that is of the model $\eta,\Sigma$.

\begin{equation}\label{eq:paramposterior}
    \begin{array}{l}
\beta(1) :=(\beta^0(1),\beta^d(1),\beta^s(1),\beta^z(1)) \sim \mathcal{N}(\eta(1),\Upsilon(1)) \\
\eta^0(1) = \eta^0(0)+\frac{1}{N}\bar{\eta}^0(1)-\Sigma(0,1)\Sigma^{-1}_{dpa(i)}(0,0)\eta^0_{dpa(i)}(0) \\
\eta^d(1) =\eta^d(0)+\Sigma^{-1}_{\setminus d}(0,0;1) \Sigma_{d}(0,1) \\
\eta^s(1) =\eta^s(0)+\Sigma^{-1}_{\setminus s}(0,0;1) \Sigma_{s}(0,1) \\
\eta^z(1) =\eta^z(0)+\Sigma^{-1}_{\setminus z}(0,0;1) \Sigma_{z}(0,1) \\
\Upsilon(1) = \psi\Upsilon(0)+\Sigma(1,1)-\Sigma(0,1)\Sigma^{-1}(0,0;1)\Sigma(0,1) \\
\Sigma(1,1) = \frac{1}{N^2}(\nu(0)-\bar{\nu}(1))(\nu(0)-\bar{\nu}(1))^T \\
\Sigma(0,1) = \frac{1}{N^2}\left(\nu(0)-\bar{\nu}(1)\right)\left(\begin{pmatrix} \sum_n\bar{X}^{(n)}_i(1) \\ \sum\bar{V}^{(n)}_{j\in dpa(i)}(1)\end{pmatrix}-\bar{\nu}(1)\right)^T \\
\Sigma(0,0;1) = \frac{1}{N^2}\left(\begin{pmatrix} \sum_n\bar{X}^{(n)}_i(1) \\ \sum\bar{V}^{(n)}_{j\in dpa(i)}(1)\end{pmatrix}-\bar{\nu}(1)\right)\left(\begin{pmatrix} \sum_n\bar{X}^{(n)}_i(1) \\ \sum\bar{V}^{(n)}_{j\in dpa(i)}(1)\end{pmatrix}-\bar{\nu}(1)\right)^T
    \end{array}
\end{equation}

Recalling that, and $W_{ab},a,b\in\{0,d,s\}$ the block mean vector and covariance matrix components corresponding to the estimates for $\beta_0,\beta_d,\beta_s$, respectively, can be similarly computed through $\Upsilon(1)$. 


We now perform the grand inductive step, to obtain the recursion from $T-1$ to $T$ to be as follows, for the posterior of the data:

\begin{equation}\label{eq:posteriordata}
\begin{array}{l}
    p\left(\mu(T) \vert W,\mathcal{S}^{(n)},\bar{\mathcal{G}}\right) \sim 
    \mathcal{N}(\nu(T),(\alpha_w+NT)W(T)),\\
    W(T)
    \sim \text{Wishart}(\alpha_w+NT,R(T)) \\
    \nu(T) = \begin{pmatrix} \mu^+(T) \\ \mu^x_{j\in dpa_d(i)}(T) \\ \mu^x_{j\in dpa_s(i)}(T) \\ \mu^z_{j\in dpa_z(i)}(T) \end{pmatrix} := \frac{1}{\alpha_{\mu}+N} \left[\alpha_{\mu}\nu(T-1)+N\bar{\nu}(T)\right]\\
    \qquad = \frac{1}{\alpha_{\mu}+N} \left[\alpha_{\mu} \nu(0)+N\sum_{t\in [T]} \bar{\nu}(t)\right] \\
    R(T) = R(T-1) + S_N(T)+\frac{\alpha_{\mu}N}{\alpha_{\mu}+N}(\nu(T-1)-\bar{\nu}(T))(\nu(T-1)-\bar{\nu}(T))^T \\
    \qquad = T+\sum\limits_{t=1}^T S_N(t)+ \frac{\alpha_{\mu}N}{\alpha_{\mu}+N}\sum\limits_{t=1}^T (\nu(t-1)-\bar{\nu}(t))(\nu(t-1)-\bar{\nu}(t))^T\\
    \bar{\nu}(T) = \begin{pmatrix} \bar{\nu}(T;T-1) \\ \bar{\nu}^x_{j\in dpa_d(i)}(T-1) \\ \bar{\nu}^x_{j\in dpa_s(i)}(T-1) \\ \bar{\nu}^z_{j\in dpa_z(i)}\end{pmatrix} := \begin{pmatrix} \frac{1}{N}\sum\limits_{n=1}^N X^{(n)}_i(T) \\ 
    \frac{1}{N}\sum\limits_{n=1}^N X^{(n)}_{j\in dpa_d}(T-1) \\ \frac{1}{N}\sum\limits_{n=1}^N X^{(n)}_{j\in dpa_s(i)}(T) \\ \frac{1}{N}\sum\limits_{n=1}^N Z^{(n)}_{j\in dpa_z(i)} \end{pmatrix} 
\end{array}
\end{equation}
In general terms, the form is broadly preserved. As such, we can reproduce the evaluation for the marginal likelihood, that is the BGe score, directly from~\cite{geiger2002parameter}

\begin{equation}\label{eq:bgscore}
\begin{array}{l}
p\left(\mathcal{S}\vert \bar{\mathcal{G}}\right) = (2\pi)^{n_x-(\tilde n_x+\tilde n_z)NT/2}\left(\frac{\alpha_{\mu}}{\alpha_{\mu}+NT}\right)^{(\tilde n_x+\tilde n_z)/2}
\frac{c(1+\tilde n_x+\tilde n_z,\alpha_{w}+(1+\tilde n_x+\tilde n_z))}{c(1+\tilde n_x+\tilde n_z,\alpha_w-1+(\tilde n_x+\tilde n_z)+NT}\\ \qquad\qquad \qquad \times 
\left\vert R(T-1)\right\vert^{\frac{a_w-n_x+\tilde n_x+\tilde n_z}{2}}\left\vert R(T)\right\vert^{-\frac{\alpha_w-n_x+\tilde n_x+\tilde n_z+NT}{2}}
\end{array}
\end{equation}
where $\tilde n_x\le n_x,\tilde n_z\le n_z$ are maximal, or the appropriate weighted average, of the sparsity of dependence on covariates on the transition to $X(t+1)$ (that is, the dimension of $V_{dpa(i)}$).

We can also express the parametric form of the posterior of the distribution of the weights, which also follows along the recursion.

\begin{equation}\label{eq:paramposterior}
    \begin{array}{l}
\beta(T) :=(\beta^0(T),\beta^d(T),\beta^s(T),\beta^z(T)) \sim \mathcal{N}(\eta(T),\Upsilon(T)) \\
\eta^0(T) = \eta^0(0)+\frac{1}{N}\bar{\eta}^0(T)-\Sigma(T-1,T)\Sigma^{-1}_{dpa(i)}(T-1,T-1)\eta^0_{dpa(i)}(T-1) \\
\eta^d(T) =\eta^d(T-1)+\Sigma^{-1}_{\setminus d}(T-1,T-1;T) \Sigma_{d}(T-1,T) \\
\eta^s(T) =\eta^s(T-1)+\Sigma^{-1}_{\setminus s}(T-1,T-1;T) \Sigma_{s}(T-1,T) \\
\eta^z(T) =\eta^z(T-1)+\Sigma^{-1}_{\setminus z}(T-1,T-1;T) \Sigma_{z}(T-1,T) \\
\Upsilon(T) = \Upsilon(T-1)+\Sigma(T,T)-\Sigma(T-1,T)\Sigma^{-1}(T-1,T-1;T)\Sigma(T-1,T) \\
\Sigma(T,T) = \frac{1}{N^2}(\nu(T-1)-\bar{\nu}(T))(\nu(T-1)-\bar{\nu}(T))^T \\
\Sigma(T-1,T) = \frac{1}{N^2}\left(\nu(T-1)-\bar{\nu}(T)\right)\left(\begin{pmatrix} \sum_n\bar{X}^{(n)}_i(T) \\ \sum\bar{V}^{(n)}_{j\in dpa(i)}(T)\end{pmatrix}-\bar{\nu}(T)\right)^T \\
\Sigma(T-1,T-1;T) = \frac{1}{N^2}\left(\begin{pmatrix} \sum_n\bar{X}^{(n)}_i(T) \\ \sum\bar{V}^{(n)}_{j\in dpa(i)}(T)\end{pmatrix}-\bar{\nu}(T)\right)\left(\begin{pmatrix} \sum_n\bar{X}^{(n)}_i(T) \\ \sum\bar{V}^{(n)}_{j\in dpa(i)}(T)\end{pmatrix}-\bar{\nu}(T)\right)^T
    \end{array}
\end{equation}

This defines the distribution of weights. Note that obtaining the score from the weights can be done straightforwardly as well, as then $W(T)$ can be recovered from $\Upsilon(T)$, and then $\mu(T)$ will have mean $\nu(T)$

\subsection{Enforcing Acyclicity}\label{s:enforcedag}
For ensuring that the structure that is learned is a proper Directed Acyclic Graph, there are a number of options as far as formulations for the various optimization problems defining the learning. Below we detail how these are enforced both when the structure is defined by integer decision variables as well as the continuous one shot formulation.

\paragraph{Integer Variables}

The primary challenge in solving optimization problems on DAGs stems from the exponential size of the acyclicity constraint. A well-known method to ensure acyclicity involves using cycle elimination constraints, which were originally introduced in the context of the Traveling Salesman Problem (TSP) in \cite{chvatal10}. Supposing that the set of all cycles is denoted by $\mathcal{C}$, these constraints often take the form
\begin{equation}
\sum_{\left(i,j\right)\in C}e_{i,j}\leq\left|C\right|-1,\quad\forall C\in\mathcal{C},
\end{equation}
where $e_{i,j}$ denote binary decision variables that indicate which edges are present in the directed graph. These constraints may be complemented by different score functions to complete the optimization problem leading to dag recovery. This can then lead to different types of problems, some of which are linear \cite{park17, jaakkola2010learning, manzour2021integer}, some quadratic \cite{rytir2024exdag}. Furthermore, this method of cycle elimination is also typically augmented with a cutting plane method \cite{Nemhauser91, rytir2024exdag}.

Another method for acyclicity enforcement is derived from a well-known combinatorial optimization problem called linear ordering (LO) \cite{gro85}. In the LO problem, we aim to find "the best" permutations, which may be further constrained. In the case a directed acyclic graphs, these permutations correspond to the placements of edges and since the basis has only quadratic cardinality, the number of constraints is limited. The cycles are then excluded by imposing LO constraints. A perceived drawback of this approach is the neccessity for a quadratic cost function \cite{seoh2020solving, gro85}.

The third method for eliminating cycles involves enforcing constraints to ensure the nodes adhere to a topological order. A topological order is a linear arrangement of the nodes in a graph such that an arc $\left(j,k\right)$ exists only if node $j$ precedes node $k$ in this order. The discrete decision variables, indexed by the node and placement in the topological order determine the graph. It has been reported that in some cases this approach can lead to polynomial time learning \cite{Shojaie10}. 

Recently, an alternative approach based on layered networks has been proposed \cite{seoh2020solving}. The concept of layering forbids the placement of arcs between layers in a given direction. The problem of finding a layered graph is defined by the number of layers and the minimal number of layers for a given DAG is unique. This contrasts the topological order method described in the previous paragraph, which can have a possitive influence on the construction of the branch-and-bound tree \cite{seoh2020solving}.

\paragraph{One Shot Continuous Formulations}
Recall that in one shot continuous variable adjacency matrix formulations, the variables denote both the structure (as far as their nonzeros) as well as the sign and magnitude of the weights themselves. Thus it is natural to consider that a constraint in the form of an equality of some function to zero could correspond to ensuring the right zero-nonzero structure of the adjacency matrix to establish acyclicity. On the other hand, considering that this must involve considerations of multiple transitions, potentially extensive matrix multiplication could, and we shall see is, involved.

The algorithm NOTEARS~\cite{zheng2018dags} and DYNOTEARS\cite{pamfil2020dynotears} uses the following functional constraint in a continuous optimization algorithm to enforce the DAG structure of the graph,
\begin{equation}\label{eq:dynoteardag}
    \mathop{tr}\exp\left\{W\odot W\right\}-d = \mathop{tr}\left(I+W+\frac{W^2}{2}+\frac{W^3}{3!}+...\right)-d= 0
\end{equation}
which is meant to approximate the following (perhaps more easily enforced) set of constraints,
\begin{equation}\label{eq:dynoteardagall}
\begin{array}{l}
    \mathop{tr}(I+W\odot W)-d = 0 \\
    \mathop{tr}(I+W\odot W\odot W)-d = 0 \\
    ...\\ 
    \mathop{tr}(I+W\odot^{n_x} W)-d = 0 
    \end{array}
\end{equation}

In \cite{yu2019dag} they introduce a different constraint term that also enforces the DAG constraint, but appears to have better numerical stability, for small $\mu>0$:
\begin{equation}
    \mathop{tr}\left((I+\mu W\odot W)^d\right)-d  
\end{equation}

In the procedure NO BEARS~\cite{lee2019scaling} the spectral radius is used to define the presence of a DAG constraint on the adjacency graph. Certain numerical approximations make this relatively feasible, despite the high complexity and nondifferentiability of the spectral radius of a matrix.

Finally, \cite{yu2021dags} present DAGS with NOCURL, which obviates the need for an explicit functional constraint by solving:
\[
(U^*,p^*) = \arg_{U\in\mathbb{S}}\min_{p\in\mathbb{R}^d} f(U \odot \mathop{ReLU}(\mathop{grad}(p)))
\]
with $\mathbb{S}$ the space of $d\times d$ skew-symmetric matrices and $\mathop{grad}(p)_{ji}=p_i-p_j$ defines the gradient flow on the nodes of the graph.

\section{Structure and Parameter Learning Algorithms}

\subsection{General Principles and Considerations}
A fundamentally unique feature of learning DBNs corresponds to how structure and weights are treated, both in and of themselves and with respect to each other, as far as modeling and training. Theoretical foundations and best practices developed in the mature disciplines of the statistics of graphical models, random graph theory, time series, causal learning, and others, can provide a diverse source of insight for developing efficient and reliable methodologies. 

Here we present a number of important points of consideration that can be observed from looking at the literature at successful attempts at representation as far as inference and learning. With the distinctions described below, we are able to properly identify and characterize existing structure-weight learners, as well as suggest extensions to fill in the natural empty places in any observed taxonomy.

\paragraph{Structure Learning}
Structure Learning is the procedure of defining $\bar{\mathcal{G}}$ from data. This is a critical aspect to learning DBNs because this defines different independence structures between the random variables. Furthermore, these graphical conditional independence structures are interpretable as far as implying causal inference and discovery. It also precedes parameter learning - the space and dimensionality of the parameters in the model itself will vary as depending on the structure of the graph connections. Of course, the quality of the resulting fit on the parameter should inform the quality of the fit of the structure, insofar as it is instrumental.

Given both the rapidly exponentially exploding complexity of considering any encoding of structure, the resulting combinatorial optimization can become difficult to solve with large variable dimension. Structure learning provides a rich source of challenging problems for combinatorics, integer programming, and other discrete applied mathematics. However, at the same time, given the relative paucity of circumstances and means by which the curse of dimensionality can be mitigated, there is a degree to which structure learning serves as a significant limitation to the overall modeling procedure. This means that often, in more challenging settings, approximate suboptimal graph structure, or using alternative modeling techniques, are used.

\paragraph{Parameter Learning}

Recall from the previous Section that there is often flexibility in the choice of the statistical model that corresponds to individual potential structures. This flexibility permits for incorporating off the shelf methods attuned for specific parametric forms. 

There are some structure solvers that define and score a structure without defining parameters. These make use of binary or Gaussian models, as defined above, for which the computation of the marginal posterior is tractable. Specifically, the posterior of the graph structure given the data is computed through an integration that treats parameters as nuisance through an integration $\int p(\{X^n_i\}\vert \theta)p(\theta\vert \mathcal{G})d\theta$. In this case, a specific set of parameters is not explicitly defined, however, it can be said that parameters are computed implicitly. Indeed, the marginal likelihood of the structure is simply the integration, over the parameter space, of the posterior distribution for the parameters. 

One can note this specific phenomenon regarding the interplay of learning structure and weights as unique to DBNs. Indeed it presents a clear tradeoff between computational ease and model faithfulness. One can also consider whether the structure of parameters are more important and significant as far as the overall modeling of the system of interest, and thus choose more or less complex models, and more or less stringent and exhaustive structure search, depending on this choice.

\paragraph{Frequentist and Bayesian}
We shall use the frequentist versus Bayesian distinction to indicate a point estimate based on the optimization of a loss function or criterion, and a probabilistic model, implemented with sampling, that obtains a posterior distribution of the structure and weights given the data, respectively.
A frequentist estimate is given as a complete specific structure encoding and a specific value for the parameters. It is generally expected, or at least sought, that the relationships that the graph identifies between various nodes representing the random variables is statistically significant. This presumption often becomes unrealistic in practice, and obtaining an appropriately scaled statistically significant entire network, that is, with all significant edges, is typically unavailable in real settings. However, since DBNs are generative rather than discriminative, this is often not a practical concern, as they are a component in an overall statistical modeling pipeline. 

The alternative of Bayesian approaches allows for modeling the full distribution of uncertainty for the model considering the data. This makes the degree of confidence in the model quantitatively transparent. Thus, for any regime of data and parameters, some density could be sampled. However, the combinatorial burden of structure learning then becomes transferred to a slow mixing time. Moreover, inference will require numerical integration, and a set of samples is less interpretable to a lay user of the model. Thus, the choice between the two is generally instrumental, that is, in accordance with the ultimate modeling goal. 

An effective and commonly used technique is to employ mixtures of a finite set of structures, see~\cite{friedman2000being}. This provides flexibility and the transparent uncertainty in the model, without having to mix through the entire combinatorial space defining possible structure. 

\paragraph{Considerations Regarding the Relationship between Structure and Parameter Learning}
It is clear that the two are not independent or orthogonal, but rather the hierarchical structure, and the discrete-continuous distinction, presents a number of possible choices as far as algorithmic options. 

For instance, consider a particular point estimate of a structure and set of parameters. However, consider that the set of parameters is close to zero, and moreover, that is so close so as to include zero in a, e.g., 95\% confidence interval. In this case, it is clear that this implies that the presence of this edge itself in the graph is suspect, that is, not implied by the data to statistical significance. 

Criteria for structure still depend on the weights, even if it's implicitly through integrating the marginal likelihood. Thus, if the weights have a poorly specified prior, or the parametric form for the model is incorrect, then this will curtail the legitimacy of the structure scoring process. 

It would be expected that a structure with a low marginal likelihood should have greater uncertainty in the parameters. 

These subtle but intuitive considerations suggest that modeling and learning with DBNs is often not an off-the-shelf straightforward use of a black box tool, but requires intuition as to the nature and mechanistic properties of the system of interest. 

\paragraph{Hierarchical and One-Shot Methods}

In general one can consider most learning methods to be hierarchical in the sense of first learning the structure, and with an amortized structure estimating or sampling the weights. The use of SEMs defined by adjacency matrices including both structure and weights simultaneously introduced what can be referred to as a one-shot approach (we remark the interestingly similar recent popularity of one-shot methods for neural architecture search that simultaneously perform parameter learning~\cite{guo2020single}).

In this case, a point estimate is obtained for both the structure and the weights simultaneously by solving an appropriate optimization problem that fits both of these as decision variables to the data. 
To this end there are two approaches we see in the literature. In~\cite{manzour2021integer}  an IP (for BNs, readily adapted to DBNs) is presented that treats the structure as binary variables encoding the activation of edge links in the graph and the parameters as separate variables, and solves the challenging nonlinear mixed IP (relaxation into conic programs was considered in~\cite{kucukyavuz2023consistent}). Alternatively, the recent work DYNOTEARS~\cite{atanackovic2024dyngfn} presented a gradient based method for solving the structure-parameter learning as a purely continuous optimization problem for weight matrices in the graph. Enforcing sparsity is done to encourage accurate ground truth graph identification. 

This presents a straightforward path to solving an optimization problem using existing toolboxes to obtain a fairly accurate point estimate of the structure and parameters. Methodologically, however, we observe that specifically, there is nothing to prevent encoding a binary variable indicating that an edge is present, and a parameter having a low magnitude to the point of zero being within the margin of error (or even being exactly zero in the IP case). These are clearly contradictory as far as the meaning of the edge.

We make one additional remark going back to hierarchical approaches. Note that one can consider that the frequentist-Bayesian distinction can be applied to present a taxonomy of methods. As a curious example, many Bayesian scoring methods, e.g. 
the IP method~\cite{BARTLETT2017258}, can be considered hybrid frequentist-Bayesian.
 This is because implicitly the grading is done with a Bayesian parameter model, but a point estimate, that is one unique structure, is returned. Methodologically, we see that the advantages of a hierarchical is an offline calculation of scoring that permits the use of simple and powerful off the shelf commercial grade IP solvers, and the disadvantage is the conceptual contradiction of applying a frequentist mindset to learning structure with Bayesian models as weights. However, one can easily mitigate this in practice by sampling from multiple structures, as weighted in frequency by their respective marginal likelihoods. Regardless, theoretically, in the asymptotic regime, consistency can still be maintained with all approaches and variations thereof, however~\cite{koller2009probabilistic}.







\subsection{Methods for Learning Structure and Parameters in DBNs}
Now we describe the details of several prominent algorithms that are used to train DBNs. These are not meant to be exhaustive, nor are they even intended to be chosen among the best performing in general. Rather, we hope to present a comprehensive variety, that is, we intend that each broad type of method that is commonly used and studied has a representative among the algorithms chosen. These algorithms use very different techniques, and treat all of the aforementioned considerations regarding learning, that is, the correspondence between structure and weights, and the distinction between points and samples and hierarchical and one shot methods. In addition, approximate (or ``local") versus exact (or ``global") methods will indicate the tradeoffs associated with seeking the best solution or seeking to find a satisficing statistical model.

\subsubsection{Constraint Based} 
Under the assumptions of causal sufficiency (no hidden confounders) and faithfulness, classical algorithms developed by Spirtes et al. \cite{causaldiscov_PC_FCI} have been proven to estimate the DAG without exhaustive enumeration of possible structures (which is impossible in interesting cases). The Peter-Clark (PC) algorithm is a method to retrieve the skeleton and directions of the edges, relying on an empirical hypothesis test of Conditional Independence (CI) for each pair of variables given a subset of other variables. It starts from a complete undirected graph and deletes sequentially edges based on these CI relations.
PCMCI \cite{pcmci} is adapted to time-series datasets and works for lagged links (causes precede effects). It operates two stages: 1) PC testing which identifies a potential set of parents with high probabilities for each variable $X_j^t$. 2) using these parents as conditions for the momentary conditional independence (MCI) to address the false positives and test all variable pairs. Statistical tests ParCorr, GPDC, and CMI are used in both steps. PCMCI+ extends PCMCI to include contemporaneous links \cite{runge2022discovering}.

This is a good representative of a method that clearly prioritizes structure, and is a statistically principled frequentist technique for identifying said structure. As such there are strong asymptotic theoretical results for this method, and it is broadly accepted to be reliable as far as identifying the ground truth. As any method prioritizing structure, however, the necessity of focusing exclusively on a discrete procedure limits the scalability of this approach.

\subsubsection{Score Based:} 

There are a number of methods that attempt to either optimize to obtain or sample from a high score of a Bayesian Criterion. We include a few of these methods due to their significant difference as far as the method of optimization/sampling.

\paragraph{Integer Programming}
The Integer Programming based~\cite{BARTLETT2017258} uses the local score (BDeu, BGe, DiscreteLL, DiscreteBIC, DiscreteAIC, GaussianLL,
GaussianBIC, GaussianAIC, GaussianL0) to optimize the network, amortizing its evaluation, thus obviating the need to compute parameters to compute the score. This algorithm was later released as GOBNILP~\cite{cussens2020gobnilp} (Globally Optimal Bayesian Network learning using Integer Linear Programming). GOBNILP finds the network with the highest score under the constraint that the underlying structure can be represented as a DAG. For every node in the graph $v$ and every possible parent set $W$, binary variable $I(W \rightarrow v)$ is created. The optimization criterion is then sum over all possible vertices and all possible parent sets, where the BDeu score (we use BDeu score here as an example) for the selected parent set of every node is considered, i.e.,
\begin{equation}
    \sum_v \sum_W I(W \rightarrow v) \cdot BDeu(v, W).
\end{equation}

The constraints are then of two types. First, each vertex needs to have a single parent set, which for node $v$ formulates as
\begin{equation}
    \sum_W I(W \rightarrow v) = 1.
\end{equation}
The second constraint requires that there are no cycles in the graph. This is imposed by \emph{cluster constraints}, which require that there must be $1$ node with no parents for any set of nodes. As there are exponentially many such sets of nodes, the optimization problem is solved, and if a cycle is in the final solution, the cluster constraint that prohibits the found cycle is added. Such computation is iterated until a DAG is found, which ensures that the optimal model is obtained.

This algorithm represents the curious ``frequentist-Bayesian" approach to structure-parameter learning. As it is an IP based method, there are also practical limitations in regards to scaling, however, the method is broadly known to be reliable and, for its search space, efficient.

\paragraph{GFlowNets}

GFlowNet (GFN) for structure learning \cite{pmlr-v180-deleu22a} consists of approximating the posterior instead of finding a single DAG, to reduce uncertainty over models. They construct the sample DAG from the posterior as a sequential decision problem by starting from an empty graph and adding one edge at a time. The GFN environment is similar to previous approaches in Reinforcement Learning, where the states are different graphs, each associated with a reward, which is in this case the score of that structure. They define a terminal state $s_f$ to which every connected state is called complete. The possible actions are limited to edge adding (no edge reversal nor removal). In addition, they define a mask that prevents having cycles in the graphs. GFN's goal is to model the whole distribution proportional to the rewards. It also borrows from Markov chain literature, using forward and backward transition probabilities, $P_{\theta}(s'/s)$ and $P_{B}(s/s')$, in the loss function that satisfies the detailed balance condition:
\begin{equation}\label{detailedbalanceloss}
     \mathcal{L}(\theta) = \sum_{s \rightarrow s^{\prime}}  \left[ log \frac{R(s^{\prime})P_{B}(s/s^{\prime})P_{B}(s_{f}/s)}{R(s)P_{\theta}(s^{\prime}/s)P_{\theta}(s_{f}/s^{\prime})} \right]^{2}
\end{equation}
where $R(s)$ is the reward function of state $s$.
Extending GFN to DBN required changing the scoring functions BDe and BGe adequately and changing the static GFN's mask to take into account the stationarity assumption (transitions are invariant in time) and present it as a block upper triangular matrix, to ensure that no edges are going from time slice $t+1$ to $t$.

This has been recently extended in~\cite{deleu2024joint} for sampling the structure and weights simultaneously using recent developments of expanding the GFN environment to continuous variables~\cite{lahlou2023theory}.

\paragraph{Monte Carlo Greedy Hill Search}
Monte Carlo methods are classical for solving difficult statistical problems, and have been a popular choice for learning the structure and parameters of a DBN. There are two prominent Monte Carlo methods in the literature that developed the foundations and have been seminal in the development of structure learning algorithms. These include the work (1128 citations as of this writing) ~\cite{friedman2000being} as well as (2344) \cite{tsamardinos2006max}, who developed the popular MMHC, a max-min hill climb (MMHC) procedure. 

In the numerical experiments, we use DMMHC from the package dbnR, which uses a variation of dynamic max-min hill climbing \cite{trabelsi2013new}, proper for leaning DBNs.

\paragraph{MCMC}
We use~\cite{kuipers2022efficient}, a more recent development. It uses order based structure sampling and at the same time restricts the search space using conditional independence tests. This method, implemented in the BiDAG package, can be viewed in a way as a hybrid based method, combining both constraint and score based methodologies.  
Performance of the method is generally the strongest for smaller datasets.  

\subsubsection{One Shot Linear SEMs:} 

There are two prominent procedures that represent one shot learning of LSEMs. The two are based on integer and continuous based optimization. LSEMs indeed uniquely present the opportunity for continuous optimization methods, and as such present the possibility of scaling the estimation procedure, at the cost of theoretical guarantees of global convergence. 

\paragraph{Integer Programming}

The mixed integer-linear program defined in~\cite{manzour2021integer} is presented here:

    \begin{equation}\label{eq:ipos}
    \begin{array}{rl}
\min\limits_{(E_W,E_A,,W,A)} & \mathbf{E}(E_W,E_A,W,A)+\lambda_W \|E_W\|_0+\lambda_A\|E_A\|_0 \\ &:=\sum\limits_{m=1}^M\sum\limits_{t=1}^T\sum\limits_{i=1}^d\left([X_{m,t}]_i-\sum\limits_{j=1}^d W_{j,i}[X_{m,t}]_j\right.\\&\left.-\sum\limits_{l=1}^{\max\{p,t\}}\sum\limits_{j=1}^n  A_{l,j,i} [X_{m,t-l}]_j\right)^2+\lambda_W \sum\limits_{i,j} [E_W]_{i,j}+\lambda_A\sum\limits_{l,i,j}[E_A]_{l,i,j} \\
\text{s.t. } &  W\cdot (1-E_W)=0,\\
& A\cdot (1-E_A)=0,\\
& DAG(E_W), \\
& (E_W,E_{A})\in \left[\{0,1\}^{d^2}\right]\times\left[\{0,1\}^{d^2}\right]^p \\
& W\in\mathbb{R}^{d\times d},\, A\in\mathbb{R}^{p\times d\times d}
    \end{array}
\end{equation}
We can see that a linear model is fit with a standard least squares loss to the data. The constraints appear, in order, as enforcing that an absent structure, defined by the binary variable $[E_W]_{i,j}=0$, corresponds to a zero weight, that is $[W]_{i,j}$, and similarly for $A$. Next, we enforce a DAG constraint on the integer variables. This was described above in the previous section. Finally, the binary and continuous variables are indicated.

\paragraph{Continuous Optimization} 

We begin by presenting the general algorithm introduced in~\cite{pamfil2020dynotears} which followed the well cited \cite{zheng2018dags}. In this paper, they consider the transition dynamics of $X(t)$ can be expressed using the SEM:
\begin{equation}\label{eq:dynotearssem}
    X_{t+1} = X_t W+ \sum\limits_{\tau=1}^{\tau_M} X_{t-\tau} A_{\tau} + \sigma_t
\end{equation}
which includes the transition encoding $W$, whose sparsity pattern reflects the patterns of causation and magnitudes the linear regression coefficients in the transition. $A_i$ are autoregressive matrices in case of lagged effects. Here $\sigma_t$ is the noise (note that in the original, this is denoted as $Z_t$, which we avoid for confusion).

They solve the optimization problem,
\begin{equation}\label{eq:dynotearsopt}
    \begin{array}{rl}
\min\limits_{W,A} & \frac{1}{2n}\sum\limits_{t,i}\|X^i_{t} - X^i_{t} W+ \sum\limits_{\tau=1}^{\tau_M} X^i_{t-\tau} A_{\tau}\|^2+\lambda_W \|W\|_1+\lambda_A\|A\|_1 \\
\text{subject to }& \mathop{Tr}\left[ \exp \left(W \circ W \right)\right]-d=0
    \end{array}
\end{equation}
wherein the nonlinear constraint function is based on the description in Section~\ref{s:enforcedag}, in particular, see the motivation by~\eqref{eq:dynoteardag}.

This method is able to impressively identify the ground truth structure for many synthetic examples, while also performing well as far as predictive modeling and forecasting of real world datasets. 

\section{Numerical Results}


The synthetic datasets were generated following causaLens \cite{lawrence2021data} and the DBN data simulation code from the DBNclass package (available at \url{https://github.com/cbg-ethz/DBNclass}) from \cite{psuter}. For causaLens, we set the maximum lag to 1 and the graph complexity to 30, representing complex causal graphs. 
\begin{itemize}
 \item \textbf{GFN}: the prior network is learned using the original GFN algorithm from \cite{pmlr-v180-deleu22a}, and we learn the transition network using the modified version of the code with the BGe score for the DBN. The hyperparameters were kept as default which can be found on the GitHub repository \url{https://github.com/tristandeleu/jax-dag-gflownet} and  we only change the number of iterations according to each experiment. 
    \item \textbf{Dynotears}: for small datasets we do a grid search for the hyperparameters over the following values: $lambda\_a$ and  $lambda\_w \in \{0.01,0.05, 0.1, 0.2,0.5\}$ and $w\_threshold \in \{0.0,0.01, 0.02, 0.05\}$. For larger datasets ($(20,400,400)$, $(30,60,100)$, $(30,600,500)$) that require more time, we only use the algorithm with hyperparameters $lambda\_w$ and $lambda\_a$ equal to $0.1$, and a small $w\_threshold$ of $0.01$.
    \item \textbf{GOBNILP}: the algorithm only supports IID data. To use it, we run it twice on the data: on the first time slice to get the prior network, then on the two first slices to get the transition network.
     \item \textbf{MCMC}: we use iterative MCMC followed by order MCMC from BiDAG package, to sample the MAP DAG. We start by a grid search for the small datasets where we choose the values $alpha \in \{0.01,0.05, 0.1, 0.2,0.3,0.4,0.5\}$ and we set the $hardlimit$ on the size of parent sets as $2n$ where $n$ is the number of covariates. For the large datasets, we restrict $alpha$ to $0.01$ and change $hardlimit$ according to the number of variables per experiment ($hardlimit=20$ for $n=20$ and $hardlimit=10$ for $n=30$).
    \item \textbf{PCMCI+}: we choose a range of $pc\_alpha \in \{0.01,0.05, 0.1, 0.2, 0.5, 0.7\}$ for small datasets, and starting from $n=20$ and $n=30$, we fix $pc\_alpha$ at 0.01. We choose ParCorr as the conditional independence test (which assumes univariate, continuous variables with linear dependencies and Gaussian noise). We further correct the p-values by False Discovery Rate control with an  $alpha\_level$ of $0.01$.
     \item \textbf{HC-dbnR}: we keep default parameters (see \url https://github.com/dkesada/dbnR).

\end{itemize}


\paragraph{Structure of Numerical Comparisons} We can consider two main purposes for which DBNs may be used, and so we perform tests comparing the learners for both criteria in an appropriate manner. In addition, we present results across the scale of small and medium covariate dimension problems. 

\begin{enumerate}
    \item \textbf{Generative Accuracy} A DBN is a generative model, meaning there are no labels, however, it is still meant to model the relationship between random variables. Thus a natural comparison as to the overall statistical quality of a model would be the classic train-test data split comparison of loss. That is, using a holdout validation set from the data, perform the learning to define a DBN model on the training data, and then perform a set of inference queries on this model, and compare their output to the ground truth output given by the validation set. 
    \item \textbf{Ground Truth Graph Identification} One of the primary goals of using BN and DBN models for fitting various time-varying phenomena is causal discovery and causal inference. This amounts to being able to accurately reconstruct the graph from a noisy realization of the ground truth. Indeed under the causal identifiability assumption given above, the relative success by which a learner is able to compute this ground truth graph is, understandably so, a central for evaluating DBN learners in the literature.
\end{enumerate}

\paragraph{Data Regimes}:
\emph{Favorable Regime for Identification}: This corresponds to $NT\gg n$, in which case, the more generally well-developed methods are able to identify the ground truth graph. We shall take:
\begin{equation}\label{eq:sizesnum1}
(n,N,T)\in \{(3,30,10),(5,50,50),(10,100,200),(20,400,400),(30,600,500)\}
\end{equation}

\emph{High Dimensional Regime}: In this case, causal identification will not be available because the number of trajectories and time steps is insufficient to specify the exact graph that generated the data. However, we can still attempt to train DBN models that fit the data appropriately. 
\begin{equation}\label{eq:sizesnum2}
(n,N,T)\in \{(3,5,10),(5,10,20),(10,20,40),(20,40,50),(30,60,100)\}
\end{equation}

\noindent
For each experiment in these data regimes, we generate 10 datasets and run the algorithms. We report the mean and standard deviation across our datasets in the following sections. For the algorithms where we performed the grid search, we keep the best results based on both SHD and AUROC equally. The chosen parameters for each experiment are listed in Table \ref{table:params_dynotears} and \ref{table:params_pcmci}. As for MCMC, the best performing parameter was $alpha = 0.01$ for all the experiments.

\subsection{Model Validation Accuracy}


For accuracy's validation, we split the time series so that the first $70\,\%$ are used for training, and the remaining $30\,\%$ are for testing. Then, we use the \texttt{dbnR} package to evaluate the log-likelihood of the test data given the predicted model. For Dynotears, only the structure was used to calculate the log-likelihood (so a binary adjacency matrix instead of the weighted one). Cycles were present in the Dynotears output networks so we used an algorithm that removed edges iteratively until cycles were no longer detected. Results are presented in Tables \ref{table:favorable_LL} and \ref{table:hd_LL}. For the error E in Table \ref{table:hd_LL} for Gobnilp, a value -Inf was obtained for one of the datasets for that experiment ((3,5,10)). In the bnlearn package (from where dbnR imports the log-likelihood function), it is explained that logLik() returns -Inf if the data have probability or density equal to zero, which typically happens if the model is singular.



\begin{table}[!h]
\begin{center}
\caption{Log-likelihood for the favorable regime. TL indicates a setting that did not finish within the time limit, and E indicates a setting that ended in an error, OOM: out of memory.}
\label{table:favorable_LL}
\setlength{\tabcolsep}{3pt}
\begin{tabular}{p{2cm}|c|c|c|c|c|}
   & \multicolumn{1}{c|}{(3,30,10)} & \multicolumn{1}{c|}{(5,50,50)}& \multicolumn{1}{c}{(10,100,200)}& \multicolumn{1}{c|}{(20,400,400)}& \multicolumn{1}{c}{(30,600,500)}\\
 \hline
 GFN & -9.96$\pm$34.31 & -207.67$\pm$195.33 & -66740.91$\pm$22342.85 & -960522.9$\pm$46428.33 &  -2542030$\pm$86678.95\\
Dynotears  &  -13.50$\pm$32.62 & -245.84$\pm$201.21 & -64746.42$\pm$21756.6 & -1056631$\pm$39963.83 &  -3369917$\pm$43506.22 \\
Gobnilp & -6.89$\pm$33.65 & -188.74$\pm$205.84 & -51572.06$\pm$17188.01 & -865201.8$\pm$9170.71 &  OOM\\
MCMC & -5.11$\pm$32.07 & -69.62$\pm$207.15 & -45776.51$\pm$15531.54 &
 -771035$\pm$786.78 &  TL \\
PCMCI+ &  -10.78$\pm$32.52 & -144.57$\pm$215.43 & -67263.58$\pm$23255.77 & TL & TL \\
HC-dbnR  & -6.46$\pm$33.40 & -79.74$\pm$207.46 & -45782.72$\pm$15511.89
 & TL &  TL \\
 \hline
\end{tabular}
\end{center}
\end{table}

\begin{table}[!h]
\begin{center}
\caption{Log-likelihood for the high dimensional regime. TL indicates a setting that did not finish within the time limit, and E indicates a setting that ended in an error, OOM: out of memory.}
\label{table:hd_LL}
\setlength{\tabcolsep}{3pt}
\begin{tabular}{p{2cm}|c|c|c|c|c|}
   & \multicolumn{1}{c|}{(3,5,10)} & \multicolumn{1}{c|}{(5,10,20)}& \multicolumn{1}{c}{(10,20,40)}& \multicolumn{1}{c}{(20,40,50)}& \multicolumn{1}{c}{(30,60,100)}\\
 \hline
 GFN & -36.89$\pm$9.98 & -341.29$\pm$17.37 & -2809.70$\pm$78.68 &  -14166.43$\pm$247.86 &  -61942.44$\pm$813.47\\
Dynotears  & -32.07$\pm$7.94 & -329.96$\pm$17.44 & -2589.79$\pm$169.73 & -13860.83$\pm$233.30 &  -66696.45$\pm$736.65 \\
Gobnilp &  E & -270.77$\pm$24.94 &  -2151.40$\pm$76.75 & -10719.46$\pm$197.12 & OOM \\
MCMC & -29.89$\pm$8.82 & -237.45$\pm$15.95 & -1773.13$\pm$34.47 & -8755.98$\pm$127.74 & TL \\
PCMCI+ & -46.28$\pm$8.94 & -363.10$\pm$14.59 & -2898.88$\pm$80.93 & -14876.04$\pm$256.19 & TL  \\
HC-dbnR  & -31.04$\pm$11.34 & -240.75$\pm$17.61 & -1819.80$\pm$38.34 & -8991.89$\pm$157.52 &  TL\\
 \hline
\end{tabular}
\end{center}
\end{table}

We see that overall it appears that PCMCI+ maintains the highest accuracy, however, for larger problem sizes becomes impractical, at which point Dynotears and GFN become the two serious options.

\newpage
\subsection{Structure Identification}

To evaluate the qualitative measures of the predicted structure, we compared the predictions with the ground truth adjacency matrix. The comparison was made using the Structural Hamming distance (SHD), which is informally the number of edges that need to be either removed from or added to the predicted structural graph to convert it to the ground truth graph. The second measure is the AUROC, a standard metric that measures the area under the receiver operator characteristic. 
For GFN, we used a 99th percentile threshold on the posterior for each experiment to only retain the highest probabilities. 
The results can be found in Tables \ref{table:favorable_roc_shd} and \ref{table:hd_roc_shd}. In PCMCI+ experiments for the high dimensional regime in Table \ref{table:hd_roc_shd}, the algorithm returns an empty matrix, which explains the 0.5 AUROC in all the experiments.  

\begin{table}[!h]
\begin{center}
\caption{Expected SHD and AUROC for favorable dimensional regime for identification. TL indicates a setting that did not finish within the time limit, and E indicates a setting that ended in an error, OOM: out of memory.}
\label{table:favorable_roc_shd}
\scriptsize
\setlength{\tabcolsep}{3pt}
\begin{tabular}{p{1.3cm}|cc|cc|cc|cc|cc|}
   & \multicolumn{2}{c|}{(3,30,10)} & \multicolumn{2}{c|}{(5,50,50)}& \multicolumn{2}{c|}{(10,100,200)}& \multicolumn{2}{c|}{(20,400,400)}& \multicolumn{2}{c}{(30,600,500)}\\
& SHD & AUROC & SHD & AUROC& SHD & AUROC& SHD & AUROC& SHD & AUROC\\
 \hline

 GFN   & 7.40$\pm$1.13 & 0.52$\pm$0.07 & 23.08$\pm$3.00 & 0.54$\pm$0.03 & 40.14$\pm$20.48 & 0.51$\pm$0.02 & 105.97$\pm$17.10 & 0.63$\pm$0.06 & 186.22$\pm$28.13 & 0.65$\pm$0.03\\
 Dynotears  &  7.13$\pm$1.29 & 0.52$\pm$0.05 & 24.40$\pm$2.80 & 0.50$\pm$0.03 & 46.56$\pm$18.04 & 0.61$\pm$0.06 & 94.34$\pm$5.59 & 0.54$\pm$0.02 & 147.20$\pm$12.22 & 0.54$\pm$0.03  \\
Gobnilp & 8.7$\pm$2.16 & 0.55$\pm$0.08 & 27.9$\pm$4.07 & 0.51$\pm$0.05 & 42.3$\pm$20.85 & 0.66$\pm$0.07 & 114.2$\pm$17.31 & 0.65$\pm$0.05 &  OOM &  OOM\\
MCMC & 7.75$\pm$2.96 & 0.61$\pm$0.10 & 16.83$\pm$4.46 & 0.70$\pm$0.08 & 98.54$\pm$10.05 & 0.56$\pm$0.07 & 391.6$\pm$15.20 & 0.59$\pm$0.06 & TL & TL\\
PCMCI+   & 6.6$\pm$1.43 & 0.59$\pm$0.03  & 19.2$\pm$2.86 & 0.61$\pm$0.04 & 58.5$\pm$18.16 &  0.53$\pm$0.02 & TL & TL & TL & TL \\
HC-dbnR  & 7.3$\pm$1.49 & 0.56$\pm$0.09  & 23.9$\pm$5.17 & 0.59$\pm$0.07 & 91.0$\pm$10.21 & 0.62$\pm$0.08 & TL & TL & TL & TL  \\
 \hline
\end{tabular}
\end{center}
\end{table}

\begin{table}[!h]
\begin{center}
\caption{Expected SHD and AUROC for high dimensional regime. TL indicates a setting that did not finish within the time limit, and E indicates a setting that ended in an error, OOM: out of memory.}
\label{table:hd_roc_shd}
\scriptsize
\setlength{\tabcolsep}{3pt}

\begin{tabular}{p{1.3cm}|cc|cc|cc|cc|cc|}
   & \multicolumn{2}{c|}{(3,5,10)} & \multicolumn{2}{c|}{(5,10,20)}& \multicolumn{2}{c|}{(10,20,40)} & \multicolumn{2}{c|}{(20,40,50)} & \multicolumn{2}{c|}{(30,60,100)} \\
& SHD & AUROC & SHD & AUROC& SHD & AUROC& SHD & AUROC& SHD & AUROC\\
 \hline
 GFN    &  8.31$\pm$1.17 & 0.56$\pm$0.09 & 15.78$\pm$3.58 & 0.51$\pm$0.04 & 33.68$\pm$7.12 & 0.53$\pm$0.02 & 76.15$\pm$8.82 & 0.54$\pm$0.02 & 134.23$\pm$13.59 & 0.54$\pm$0.02 \\
Dynotears  &  8.49$\pm$1.29 & 0.69$\pm$0.08 & 17.02$\pm$3.00 & 0.64$\pm$0.10 & 38.67$\pm$5.06 & 0.62$\pm$0.04 & 86.66$\pm$9.30 & 0.63$\pm$0.03 & 143.44$\pm$9.90 & 0.54$\pm$0.03 \\
Gobnilp   & 8.0$\pm$2.62 & 0.63$\pm$0.10 & 17.2$\pm$4.71 & 0.60$\pm$0.09 & 46.4$\pm$9.83 & 0.60$\pm$0.06 & 96.1$\pm$9.40 & 0.65$\pm$0.04 & OOM & OOM \\
MCMC   & 10.00$\pm$2.26 & 0.53$\pm$0.09 & 21.86$\pm$6.14 & 0.59$\pm$0.10 & 79.72$\pm$7.54 & 0.58$\pm$0.05  & 264.5$\pm$15.04 & 0.59$\pm$0.05  & TL & TL\\
PCMCI+   &  8.2$\pm$1.03 & 0.5$\pm$0.0 & 15.0$\pm$3.65 & 0.5$\pm$0.0 & 32.5$\pm$6.04 & 0.5$\pm$0.00 & 67.4$\pm$10.12& 0.50$\pm$0.00  & TL & TL\\
HC-dbnR  & 7.1$\pm$3.38 & 0.64$\pm$0.14 & 17.6$\pm$6.11 & 0.65$\pm$0.11 & 68.0$\pm$5.68 & 0.63$\pm$0.05 & 233.4$\pm$19.46 & 0.60$\pm$0.04  & TL & TL\\
 \hline
\end{tabular}
\end{center}
\end{table}

\begin{table}[!h]
\begin{center}
\caption{Dynotears parameters.}
\label{table:params_dynotears}
\setlength{\tabcolsep}{3pt}
\begin{tabular}{p{2cm}|c|}
   & \multicolumn{1}{|c|}{($lambda\_w$,$lambda\_a$,$w\_threshold$)} \\
 \hline
 (3,30,10) & (0.5,0.2,0.01) \\
 (5,50,50) & (0.5,0.2,0.05)\\
(10,100,200) & (0.01,0.2,0.0)  \\
(20,400,400) & (0.1,0.1,0.01) \\
(30,600,500) &  (0.1,0.1,0.01)\\
 (3,5,10) & (0.2,0.1,0.0)\\
  (5,10,20)  & (0.1,0.05,0.0)\\
(10,20,40) & (0.2,0.05,0.0)\\
(20,40,50) & (0.2,0.01,0.0)\\
(30,60,100) & (0.1,0.1,0.01)\\
 \hline
\end{tabular}
\end{center}
\end{table}

\begin{table}[!h]
\begin{center}
\caption{PCMCI+ parameter.}
\label{table:params_pcmci}
\setlength{\tabcolsep}{3pt}
\begin{tabular}{p{2cm}|c|}
   & \multicolumn{1}{|c|}{pc\_alpha} \\
 \hline
 (3,30,10) &  0.2\\
 (5,50,50) & 0.05\\
(10,100,200) & 0.7 \\
 (3,5,10) & 0.01\\
  (5,10,20)  & 0.01\\
(10,20,40) & 0.01\\
(20,40,50) & 0.01\\
 \hline
\end{tabular}
\end{center}
\end{table}

We see that again for smaller dimensions, PCMCI+ appears to be the most reliable (although by a small margin). For larger problems, again GFN and Dynotears become the best performers, with a slight edge towards GFN. 

\newpage

\section{Discussion and Conclusion}
We hope this paper has provided a useful guide to the main principles behind learning the structure and parameters of a DBN. We focused on the fundamentals for the most simple cases, while targeting breadth in the scope of the various methodological approaches to learning these models from data. 

There is an important aspect to DBNs that we did not discuss, as for the simple cases of learning it can be considered an orthogonal topic. This would be inference. DBNs are a generative model, so by themselves they do not accomplish any particular statistical decision test. However, one can perform various inference queries, such as the probability that an instance of $X(2)$, given $X(1)=3.2$ and $Z=3$, would be greater than $2.1$. A natural query for DBNs is a forward time forecast. Causal inference can also be performed through queries using DBN models. Inference and approximate inference have a number of different procedures available, as far as efficiently and effectively sampling from the network. 

Furthermore, inference algorithms are required in order to further extend DBN modeling to many real world datasets. For one, they become necessary for the expectation step in an Expectation-Maximization algorithm to learn structure with hidden variables. Often, with systems wherein the mechanism of action isn't observed, a latent variable structure is able to model a rough set of possible dependencies that fits the observed data directed to and from in the graph. 

For larger dimensions, IP approaches become computationally infeasible. In such a circumstance, given the Sample Complexity discussed in Section~\ref{s:learningfromtraj}. 

When data is plentiful, that is, millions and possibly easily available streaming samples, then neural network approaches can be effective. This suggests, for instance, the potential scalability of Generative Flow Networks~\cite{atanackovic2024dyngfn} for instance. Reinforcement Learning is another common approach~\cite{zheng2023rbnets}. Otherwise, in the high-dimensional regime, wherein samples are finite but there are many covariates, Bayesian methods~\cite{besada2009parallel} or meta-heuristics are typically applied~\cite{hu2011integrated}.


\section*{Acknowledgements} 
The authors would like to thank Ond{\v r}ej Ku{\v z}elka for his suggestions and discussion on this work. This work has received funding from the European Union’s Horizon Europe research and innovation programme under grant agreement No. 101084642.
\bibliographystyle{plain}
\bibliography{refs}
\end{document}